\documentclass{article}

 \PassOptionsToPackage{numbers, compress}{natbib}

\usepackage[preprint]{neurips_2022}

\usepackage{url}            
\usepackage{booktabs}       
\usepackage{nicefrac}       
\usepackage{microtype}      
\usepackage{adjustbox}
\usepackage[flushleft]{threeparttable}
\usepackage{wrapfig}
\usepackage{tcolorbox}
\usepackage[page, header]{appendix}
\usepackage[utf8]{inputenc} 
\usepackage[T1]{fontenc}    
\usepackage{url}            
\usepackage{booktabs}       
\usepackage{caption}
\usepackage{amsfonts}       
\usepackage{nicefrac}       
\usepackage{microtype}      
\usepackage{bm}
\usepackage{diagbox}
\usepackage{color} 
\usepackage{algorithmic}
\usepackage{amssymb}
\usepackage{amsmath}
\usepackage{amsfonts}
\usepackage{enumitem}

\usepackage{mathtools}
\usepackage{lineno}
\usepackage{minitoc}
\doparttoc 
\faketableofcontents 
\usepackage[labelformat=simple]{subcaption}
\usepackage{wrapfig}
\newenvironment{smalleralign}[1][\small]
 {\par\nopagebreak\leavevmode\vspace*{-\baselineskip}%
  \skip0=\abovedisplayskip
  #1%
  \def\maketag@@@##1{\hbox{\m@th\normalfont\normalsize##1}}%
  \abovedisplayskip=\skip0
  \align}
 {\endalign\ignorespacesafterend}
\makeatother
\newtheorem{proof}{Proof}

\newtheorem{theorem}{Theorem}
\newtheorem{definition}{Definition}
\newtheorem{proposition}{Proposition}
\newtheorem{corollary}{Corollary}
\newtheorem{lemma}{Lemma}
\newcommand{\switcher}{{\fontfamily{cmss}\selectfont Switcher}}
\newcommand{\exploiter}{{\fontfamily{cmss}\selectfont Exploiter}}
\newcommand{\exploit}{{\,}} 
\newcommand{\explore}{{\rm xplr}}

\usepackage{natbib}
\usepackage{multicol, multirow}
\usepackage[ruled,vlined]{algorithm2e}

\newcommand\blfootnote[1]{%
\begingroup
\renewcommand\thefootnote{}\footnote{#1}%
\addtocounter{footnote}{-1}%
\endgroup
}

\newcommand{\cR}{{\mathcal R}} 
\newcommand{\cS}{{\mathcal S}}

\SetKwInput{KwInput}{Input}                
\SetKwInput{KwOutput}{Output} 
\newtheorem{innercustomthm}{Theorem}
\newenvironment{customthm}[1]
  {\renewcommand\theinnercustomthm{#1}\innercustomthm}
  {\endinnercustomthm}
\providecommand{\customgenericname}{}
\newcommand{\newcustomtheorem}[2]{%
  \newenvironment{#1}[1]
  {%
  \renewcommand\customgenericname{#2}%
  \renewcommand\theinnercustomgeneric{##1}%
  \innercustomgeneric
  }
  {\endinnercustomgeneric}
}
\newcustomtheorem{customtheorem}{Theorem}

\usepackage{hyperref}
\hypersetup{
    colorlinks=true,
    linkcolor=red,
    citecolor=blue,
    filecolor=magenta,      
    urlcolor=blue,
linktocpage}
\usepackage{cleveref}

\title{SEREN: Knowing When to Explore and \\When to Exploit}

\author{
Changmin Yu$^{*, \dag}$ \textsuperscript{1}
David Mguni$^{*,\dag}$\textsuperscript{2},
Dong Li\textsuperscript{2}, 
Aivar Sootla\textsuperscript{2},
Jun Wang\textsuperscript{1, 2},
Neil Burgess\textsuperscript{1}
\\
\textsuperscript{1}UCL, London, United Kingdom\\
\textsuperscript{2}Huawei Noah's Ark Lab

}
  
\begin{document}
\maketitle

\begin{abstract}

Efficient \blfootnote{$^*$Equal contribution.  $^\dag$Corresponding author  <changmin.yu.19@ucl.ac.uk> <david.mguni@hotmail.com>. }reinforcement learning (RL) involves a trade-off between ``exploitative" actions that maximise expected reward and ``explorative'" ones that sample unvisited states. To encourage exploration, recent approaches proposed adding stochasticity to actions, separating exploration and exploitation phases, or equating reduction in uncertainty with reward. However, these techniques do not necessarily offer entirely systematic approaches making this trade-off. Here we introduce \textbf{SE}lective \textbf{R}einforcement \textbf{E}xploration \textbf{N}etwork (SEREN) that poses the exploration-exploitation trade-off as a game between an RL agent--- \exploiter, which purely exploits known rewards, and another RL agent--- \switcher, which chooses at which states to activate a \textit{pure exploration} policy 
that is trained to minimise system uncertainty and
override {\fontfamily{cmss}\selectfont Exploiter}. Using a form of policies known as \textit{impulse control}, \switcher~is able to determine the best set of states to switch to the exploration policy while {\fontfamily{cmss}\selectfont Exploiter} is free to execute its actions everywhere else. We prove that SEREN converges quickly and induces a natural schedule towards pure exploitation. Through extensive empirical studies in both discrete (MiniGrid) and continuous (MuJoCo) control benchmarks, we show that SEREN can be readily combined with existing RL algorithms to yield significant improvement in performance relative to state-of-the-art algorithms.

\end{abstract}




\section{Introduction}
Reinforcement learning (RL) is a framework that enables autonomous agents to learn complex behaviours without human intervention \cite{sutton2018reinforcement}. 
RL has had notable successes in a number of practical domains such as robotics and video games \cite{deisenroth2011learning,peng2017multiagent}. During the training phase, an RL agent learns about the value for each state using a \textit{trial-and-error approach} to determine the best actions across the state space. 
For the agent to obtain a sufficient coverage of the state space for finding the globally optimal policy, actions must include stochasticity as well as ``greed" (i.e. maximizing current expected reward) ~\cite{sutton2018reinforcement}.
However, randomly perturbing actions is sample inefficient since the it does not take into account the environment information previous experiences.  
In practice, this procedure exacerbates the sample complexity of the agent's problem, leading to many samples being needed for learning its optimal policy, despite theoretically grounded asymptotic convergence. 
%
%

In this paper, we tackle the challenge of performing systematic and efficient exploration in RL. We propose a novel two-agent framework that disentangles the exploration and exploitation for more efficient independent learning.
We propose SElective Reinforcement Exploration Network (SEREN), which entails an interdependent interaction between an RL agent, {\fontfamily{cmss}\selectfont Exploiter}, whose goal is to maximise 
the current estimate of future rewards (either model-free or model-based)  and an additional RL agent, {\fontfamily{cmss}\selectfont Switcher}, whose goal is to
explore so as to reduce uncertainty in the Exploiter.
Furthermore, \switcher~has the power to override the \exploiter~and \textit{assume} control of the system  to apply an exploratory action. 
The individual goals of completing the task set by the environment and exploration are decoupled and each delegated to an individual agent. 

\textbf{Why use a Two-Agent Framework?}

%
%

\textbf{Selective exploration.} Key to SEREN is the introduction of another agent which learns the best set of states to perform exploration given the current measure of uncertainty.  Moreover, as we formally prove in Sec \ref{sec:theory}, a schedule of exploration naturally emerges from SEREN which does not require heuristic exploration scheduling (see Prop. \ref{prop:switching_times}). Additionally, techniques such as $\epsilon$-greedy policies can be readily seen to be a degenerate case of our framework. 

\textbf{Decoupled objectives \& exploration planning.} The tasks of maximising the environmental reward and minimising uncertainty about unexplored states are \textit{fully decoupled}. This means {\fontfamily{cmss}\selectfont Exploiter} pursues its task of maximising its objective without the necessity of trading-off exploratory actions for environmental rewards.




\textbf{Plug \& play.} 
SEREN is a general framework that can be instantiated with different RL base learner, and can flexibly accommodate different exploration objective. We empirically show this in Sec.~\ref{sec:experiments}.
%

\textbf{Neuroscience/behavioural correspondence.}
A well-established hypothesis of animal decision making is that animals exhibit information-seeking behaviour to reduce internal estimate of the uncertainty of the environment~\cite{gottlieb2013information}. Experimental evidence indicates the orthogonal encoding of information value and primary reward value in primate orbitofrontal cortex (OFC) for curiosity-based decision making~\cite{blanchard2015orbitofrontal}, which coheres nicely with SEREN's framework of dual system for independent learning of exploitative and exploratory behaviours. See further discussion in Section~\ref{Section:Conclusion}. 
\vspace{-10pt}
%
%
%
%
%
%
%
%
%

%
%
%
%

\section{Related Work}
\vspace{-5pt}
\textbf{Exploration-Exploitation Tradeoff} is a fundamental question in RL research, 
i.e. trading off finding higher reward states and exploiting known rewards. A simple but prevalent approach is directly injecting pure noise or certain parametric stochasticity into action choices during learning, such as $\epsilon$-greedy algorithm~\cite{sutton2018reinforcement}; adding noise parametrised by Ornstein–Uhlenbeck processes~\cite{lillicrap2015continuous}; using stochastic controllers regularised by the maximum entropy principle~\cite{haarnoja2018soft}, etc. Despite the simplicity, this method does not account for system uncertainty or known rewards and therefore lacks both efficiency and interpretability. Among more systematic approaches that originate in the multi-armed bandits literature, is exploration according to `Optimism in the Face of Uncertainty' (OFU). Some popular algorithms 
under the OFU framework include the Upper Confidence Bound (UCB)  algorithm \citet{auer2002using} that achieves theoretically justified regret bounds and active inference algorithms that relate exploration with free energy maximisation under the variational inference principle~\cite{schwartenbeck2013exploration}.

\textbf{Reward free exploration}~\cite{jin2020reward}, also known as the task-agnostic or reward-agnostic setting~\cite{zhang2020task} is a closely related method. In this setting, the agent goes through a two-stage process. In the exploration phase the agent interacts with the environment without the guidance of any reward information, and in the planning phase the reward information is revealed and the agent computes a policy based on the transition information collected in the exploration phase and the reward information revealed in the planning phase. We also separate the tasks of exploration from exploitation using two processes, however we note that here the two processes are performed concurrently and actions are chosen based on either process interchangeably, hence achieving a more self-contingent tradeoff between exploration and exploitation.


\textbf{Curiosity-driven Exploration} is another popular framework for exploration, in which a prediction model is employed to measure agent's knowledge about the environment. The discrepancy between the internal estimate and the ground-truth observation is treated as intrinsic reward, which is combined with extrinsic reward to guide policy learning. \citet{Stadie2015IncentivizingEI} considers the difference between prediction of a forward dynamics model and observed state as the intrinsic reward. 
\citet{pathak2017curiosity} proposed to generate the intrinsic curiosity-based reward using an inverse dynamics model, and is able generate reward signals only correlated with the controllable aspects of the environment. However, existing methods couple the exploration and exploitation, and so may cause learning inefficiency for the exploiter.

\textbf{Uncertainty quantification in exploration}
%
is an active field of research in RL.
It is common practice to use the disagreement of the predictions over an ensemble of neural networks as the epistemic uncertainty to guide exploration~\cite{osband2016deep, janner2019trust, sekar2020planning, lee2021sunrise}. Connections between ensemble disagreement and information theory have been drawn such that choosing actions that maximises the expected ensemble disagreement would maximally increase the information gain, improving the efficiency of exploration~\cite{sekar2020planning, o2021variational}.
Other popular alternatives involve the prediction error of a discriminative dynamics model~\cite{schmidhuber1997shifting, pathak2017curiosity} and the predictive uncertainty given a generative dynamics model~\cite{ratzlaff2020implicit, jiang2020generative}.

\vspace{-5pt}
\section{Preliminaries}
\vspace{-5pt}
%

\textbf{Reinforcement Learning (RL).} In RL, an agent sequentially selects actions to maximise its expected returns. The underlying problem is typically formalised as an MDP $\left\langle \mathcal{S},\mathcal{A},P,R,\gamma\right\rangle$ where $\mathcal{S}\subset \mathbb{R}^p$ is the set of states, $\mathcal{A}\subset \mathbb{R}^k$ is the set of actions, $P:\mathcal{S} \times \mathcal{A} \times \mathcal{S} \rightarrow [0, 1]$ is a transition probability function describing the system's dynamics, $R: \mathcal{S} \times \mathcal{A} \rightarrow \mathbb{R}$ is the reward function measuring the agent's performance and the factor $\gamma \in [0, 1)$ specifies the degree to which the agent's rewards are discounted over time \cite{sutton2018reinforcement}. At time $t\in 0,1,\ldots, $ the system is in state $s_{t} \in \mathcal{S}$ and the agent must
choose an action $a_{t} \in \mathcal{A}$ which transitions the system to a new state 
$s_{t+1} \sim P(\cdot|s_{t}, a_{t})$ and produces a reward $R(s_t, a_t)$. A policy $\pi: \mathcal{S} \times \mathcal{A} \rightarrow [0,1]$ is a probability distribution over state-action pairs where $\pi(a|s)$ represents the probability of selecting action $a\in\mathcal{A}$ in state $s\in\mathcal{S}$. The goal of an RL agent is to
find a policy $\hat{\pi}\in\Pi$ that maximises its expected returns given by the value function: $
v^{\pi}(s)=\mathbb{E}[\sum_{t=0}^\infty \gamma^tR(s_t,a_t)|a_t\sim\pi(\cdot|s_t)]$ where $\Pi$ is the agent's policy set. 




%
%

\section{SEREN: Dual System Framework for Exploration-Exploitation Tradeoff} 
\label{sec: method}

\begin{wrapfigure}{r}{.45\linewidth}
\vspace{-10pt}
    \centering
    \includegraphics[width=\linewidth]{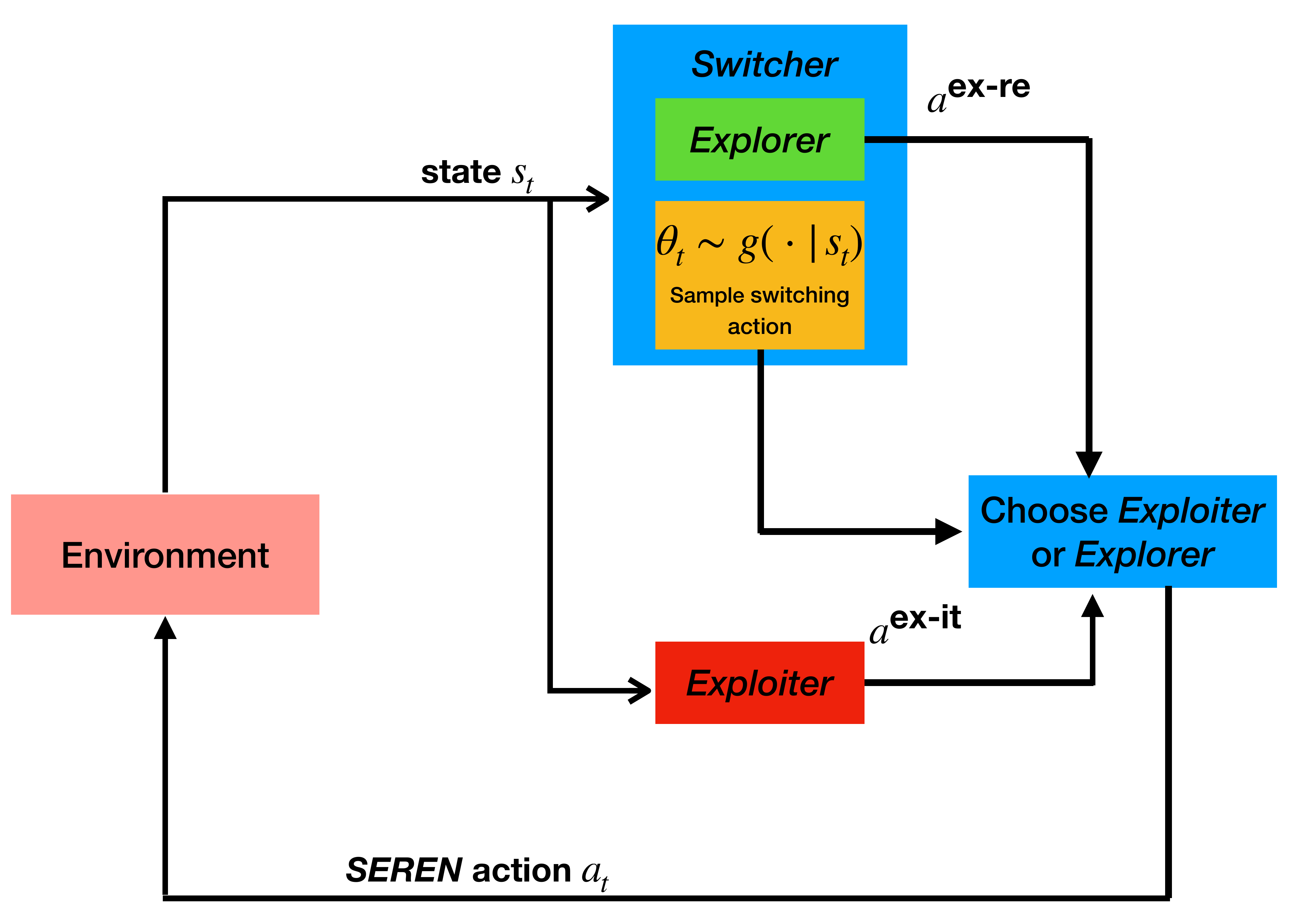}
    \caption{Schematic of the SEREN framework.}
    \label{fig: seren}
    \vspace{-15pt}
\end{wrapfigure}
Our framework, SEREN consists of an RL agent, {\fontfamily{cmss}\selectfont Exploiter} and an \textit{impulse control} agent {\fontfamily{cmss}\selectfont Switcher}. The {\fontfamily{cmss}\selectfont Switcher} has the ability to transfer control of the system to the exploration policy and does so at a set of states it chooses. As we later describe in detail, the {\fontfamily{cmss}\selectfont Switcher} uses a type of policy known as impulse control \cite{mguni2018optimal}\footnote{Our setup is related to stochastic differential games with impulse control \cite{mguni2018viscosity,cosso2013stochastic}. However, our Markov Game (MG) differs markedly since it is nonzero-sum, an agent \textit{assumes control} and is a discrete-time treatment.} which enables the {\fontfamily{cmss}\selectfont Switcher} to determine the best set of states to transfer control to the exploration policy. This means that exploration is performed only at states and conditions in which exploratory actions sufficiently reduce the system uncertainty while the {\fontfamily{cmss}\selectfont Exploiter} is free to exploit everywhere else.

Unlike standard RL in which the goal of exploration and exploitation are housed within one objective, the goal of minimising uncertainty about unexplored states and exploiting known rewards are now decoupled.

Formally, our framework is defined by a tuple $\mathcal{G}=\langle \mathcal{N},\mathcal{S},\mathcal{A}^\exploit,\mathcal{A}^\explore,P,R^\exploit,R^\explore\rangle$ where the new elements are the set of agents $\mathcal{N}=\{\text{\fontfamily{cmss}\selectfont Exploiter}, \text{\fontfamily{cmss}\selectfont Switcher}\}$, $\mathcal{A}^\exploit$ and $\mathcal{A}^\explore\subseteq \mathcal{A}$ are the exploitation (for {\fontfamily{cmss}\selectfont Exploiter}) and exploration (for {\fontfamily{cmss}\selectfont Switcher}) action sets, respectively, and the functions $R^\exploit, R^\explore:\mathcal{S}\times\mathcal{A}^\exploit\times\mathcal{A}^\explore\to\mathbb{R}$ are the one-step rewards . The transition probability $P:\mathcal{S}\times\mathcal{A}^\exploit\times\mathcal{A}^\explore\times\mathcal{S}\to[0,1]$ takes the state and action of both agents as inputs. The {\fontfamily{cmss}\selectfont Exploiter} agent has a Markov policy $\pi^\exploit:\mathcal{S}\rightarrow\mathcal{A}^\exploit$, which is contained 
in the set $\Pi^\exploit\subseteq\Pi$. 
The {\fontfamily{cmss}\selectfont Switcher} agent has two components a Markov policy $\pi^\explore:\mathcal{S}\rightarrow\mathcal{A}^\explore$ from $\Pi^\explore\subseteq\Pi$, which determines the exploration action based on a measure of uncertainty, and a (categorical) policy $\mathfrak{g}:\mathcal{S} \to \{0,1\}$, which determines when to activate exploration. 
At each state the {\fontfamily{cmss}\selectfont Switcher} makes a \textit{binary decision} to decide whether to transfer control of the system to the exploration policy $\pi^\explore$. We denote by $\{\tau_k\}_{k\geq 0}$ the points at which the {\fontfamily{cmss}\selectfont Switcher} decides to activate the exploration policy or the \textit{intervention times}, so for example if the {\fontfamily{cmss}\selectfont Switcher} chooses to switch to exploration at state $s_6$ and again at state $s_8$, then $\tau_1=6$ and $\tau_2=8$. The intervention times obey the expression $\tau_k=\inf\{t>\tau_{k-1}|s_t\in\mathcal{S},\mathfrak{g}(s_t)=1\}$ and are therefore \textit{ \textbf{rules} that depend on the state.} Hence, by learning an optimal $\mathfrak{g}$, {\fontfamily{cmss}\selectfont Switcher} learns the best states to activate exploration. As we later explain, these intervention times are determined by a condition on the state which is easy to evaluate (see Prop. \ref{prop:switching_times}). We assume that all the policies $\pi^\exploit$, $\pi^\explore$ and $\mathfrak{g}$ are computed using value function based or actor-critic based frameworks. 

\textbf{The Exploiter Objective} 

The goal of {\fontfamily{cmss}\selectfont Exploiter} is to (greedily) maximise its expected cumulative reward set by the environment. 
The objective that {\fontfamily{cmss}\selectfont Exploiter} seeks to maximise is:
\begin{smalleralign}
&v^{\pi^\exploit,(\pi^\explore,\mathfrak{g})}_1(s)=\mathbb{E}\left[\sum_{t\geq 0}\gamma_{1}^tR^\exploit\left(s_t,a^\exploit_t,a^\explore_t\right)\Big|s_0\equiv s\right],   \label{p1objective}  \end{smalleralign}
where $a^\exploit_t\sim\pi^\exploit(\cdot|s_t)$ is {\fontfamily{cmss}\selectfont Exploiter}'s action and $a^\explore_t\sim\pi^\explore$ is an action chosen according to the exploration policy, the reward function is defined by $R^\exploit(s_t,a^\exploit_t,a^\explore_t)=
R(s_t,a_t^\exploit)(1-\boldsymbol{1}_{\mathcal{A}^\explore}(a^\explore_t)) + R(s,a^\explore_t)\boldsymbol{1}_{\mathcal{A}^\explore}(a^\explore_t)$ and $\boldsymbol{1}_{\mathcal{Y}}(y)$ is the indicator function which is $1$ whenever $y\in\mathcal{Y}$ and $0$ otherwise. 
Therefore, the reward received by {\fontfamily{cmss}\selectfont Exploiter} is $R(s_t,a^\explore_{t})$ when $t=\tau_k$, $k=1,2,\ldots$ i.e. whenever the {\fontfamily{cmss}\selectfont Switcher} activates the exploration policy and $R(s_t,a_{t}^\exploit)$ otherwise.

Whenever {\fontfamily{cmss}\selectfont Switcher} decides to transfer control to the exploration policy, the exploration policy overrides the {\fontfamily{cmss}\selectfont Exploiter} and the transition dynamics are affected by only the exploration policy (while {\fontfamily{cmss}\selectfont Exploiter} influences the dynamics at all other times). 
The transition dynamics are therefore given by $\boldsymbol{P}(s_{t+1}, a^\explore_t,a^\exploit_t,s_t):=
P(s_{t+1},a^\exploit_t,s_t)\left(1-\boldsymbol{1}_{\mathcal{A}^\explore}(a^\explore_t)\right)+P(s_{t+1},a^\explore_t,s_t)\boldsymbol{1}_{\mathcal{A}^\explore}(a^\explore_t)$.
Therefore, the transition function is $P(s_{t+1},a^\explore_{t},s_t)$ when $t=\tau_k$, $k=1,2,\ldots$ i.e. whenever the {\fontfamily{cmss}\selectfont Switch} activates the exploration policy and $P(s_{t+1},a^\exploit_{t},s_t)$ otherwise.
\textbf{The Exploration Policy}  

The actions selected by the exploration policy $\pi^\explore$ are chosen so as to maximise the following:
\begin{smalleralign} \label{p2objective}
&\hat v^{\pi^\exploit,(\pi^\explore,\mathfrak{g})}_2(s)=\mathbb{E}\left[\sum_{t\geq 0}\gamma_2^t R^\explore\left(s_t,a^\exploit_t,a^\explore_t\right)\Big|s_0\equiv s\right], 
\end{smalleralign}

where  $
R^\explore(s_t,a^\exploit_t,a^\explore_t):=-\Big(L(s,a^\explore_t)\boldsymbol{1}_{\mathcal{A}^\explore}(a^\explore_t)+L(s_t,a^\exploit_{t})(1-\boldsymbol{1}_{\mathcal{A}^\explore}(a^\explore_t))\Big)$,
and $L$ is the measure of uncertainty which we specify in detail shortly which is chosen to satisfy the property that $L\to 0$ as the system uncertainty decreases. Analogous to the reward function for the {\fontfamily{cmss}\selectfont Exploiter}, the function $R^\explore$ is defined so that the received reward is $-L(s_t,a^\explore_{t})$ when $t=\tau_k$, $k=0,1,\ldots$ i.e. whenever the {\fontfamily{cmss}\selectfont Switcher} activates the exploration policy and $-L(s_t,a^\exploit_t)$ otherwise. As the learning process progresses, the uncertainty inevitably decreases making the process of learning the exploration policy non-stationary. In order to counteract the negative impacts brought by the non-stationarity of the reward structure of \switcher, the discounting factor $\gamma_{2}$ is set to be small such that the agent avoids taking the rewards of distant future states into consideration, where large degree of mismatch between the cached and current uncertainty estimation exists. 


In general, SEREN accommodates various measures of uncertainty, for instance, in the model-based case we can use epistemic uncertainty based on the ensemble of dynamical models \cite{chua2018deep} (see Appendix~\ref{app:uncertainty-measures}).
In the current setting, we focus on the model-free version of the method of uncertainty measurement. In this case, we assume that the {\fontfamily{cmss}\selectfont Exploiter} uses an ensemble of neural networks as its critic (value function) estimate. We quantify the uncertainty over the state space using a non-parametric estimate based on ensemble modelling of the value function of the {\fontfamily{cmss}\selectfont Exploiter}. As with value-based methods~\cite{osband2016deep}, we quantify the uncertainty using the epistemic uncertainty across the ensemble networks. In particular, for an ensemble of $E$ critic estimates of $\{\mathcal{Q}_{1}, \dots, \mathcal{Q}_{E}\}$,
we have the following measure of uncertainty for any $a \in\mathcal{A}$ and for any $s\in\mathcal{S}$:\looseness=-1
    \begin{smalleralign}
        L(s, a)=  \frac{1}{E-1}\sum_{e}(\mathcal{Q}_{e}(s, a) - \mu(s, a))^2,
        \label{eq:q-ensemble-var}
    \end{smalleralign}
where $\mu(s, a) := \frac{1}{E}\sum_{e}\mathcal{Q}_{e}(s, a)$ is the empirical mean of the ensemble predictions.     



\textbf{The Switcher Mechanism} 

We now describe the {\fontfamily{cmss}\selectfont Switcher}'s objective. 
The goal of the {\fontfamily{cmss}\selectfont Switcher} is to minimise uncertainty about unexplored states. 
To induce {\fontfamily{cmss}\selectfont Switcher} to selectively choose when to switch on exploration, each switch activation  incurs a fixed cost for {\fontfamily{cmss}\selectfont Switcher}. These costs are quantified by the indicator function which is $1$ whenever an exploratory action is performed and $0$ otherwise. The presence of this cost ensures that the gain for {\fontfamily{cmss}\selectfont Exploration} for performing an exploratory action to arrive at a given set of states is sufficiently high to merit forgoing rewards from exploitative actions.  Therefore to maximise its objective, the {\fontfamily{cmss}\selectfont Switcher} must determine the sequence of points $\{\tau_k\}$ at which the benefit of performing a precise action overcomes the cost of doing so. Accordingly at time $t \in 0,1,\ldots$, the {\fontfamily{cmss}\selectfont Switcher} seeks to maximise the following quantity: 
\begin{smalleralign} 
&v^{\pi^\exploit,(\pi^\explore,\mathfrak{g})}_2(s)
=\mathbb{E}\left[\sum_{t\geq 0}\gamma^t \left(R^\explore\left(s_t,a^\exploit_t,a^\explore_t\right)  - \beta\cdot\boldsymbol{1}_{\mathcal{A}^\explore}(a^\explore_t)\right)\right],    \label{p3objective}
\end{smalleralign}
where $\beta$ is some pre-specified scalar intervention cost. Therefore to maximise its objective, the {\fontfamily{cmss}\selectfont Switcher} must determine the best set of states to perform exploration, that is the set of states that reduce system uncertainty. Note that since $R^\explore$ depends on the uncertainty measure $L$, it has the property that $R^\explore\to 0$ as system uncertainty decreases.  
With low level of uncertainty $L$, the cost of switching dominates so that {\fontfamily{cmss}\selectfont Switcher} does not intervene leaving {\fontfamily{cmss}\selectfont Exploiter} to take actions that deliver high rewards. This effectively pushes {\fontfamily{cmss}\selectfont Switcher} out of the game as systemic uncertainty is reduced. This is precisely the behaviour that we seek as more about the system becomes known. We later formally prove this property of our framework in Sec. \ref{sec:theory} (see Prop. \ref{prop:switching_times}).  Note that both agents use deterministic policies so that the system naturally evolves towards full exploitation with no exploratory actions.
\vspace{-5pt}

\subsection*{Switcher Learns Faster than Exploiter}
The impulse control mechanism results in a framework in which the problem facing {\fontfamily{cmss}\selectfont Switcher} has a decision space of $\mathcal{S}\times\{0,1\}$ i.e at each state it makes a binary decision (this differs from {\fontfamily{cmss}\selectfont Exploiter} though both agents share the same experiences).  Consequently, the learning process for $\mathfrak{g}$ is relative quick (and unlike {\fontfamily{cmss}\selectfont Exploiter}'s who must optimise over a decision space which is $|\mathcal{S}|\cdot|\mathcal{A}|$, choosing an action from its action space at every state). This results in the {\fontfamily{cmss}\selectfont Switcher} rapidly learning its optimal policy, enabling it 
to efficiently guide exploration during training.
\vspace{-5pt}

\vspace{-5pt}
\subsection*{Relation to Other Exploratory Mechanisms} 
\vspace{-5pt}
Many existing exploration models can be viewed as some degenerate form of SEREN. For instance, the classical $\epsilon$-greedy exploration can be interpreted as SEREN with a random switching mechanism and uniform exploration policy. If we consider the case in which {\fontfamily{cmss}\selectfont Switcher} has an identical objective to {\fontfamily{cmss}\selectfont Exploiter}, then the model is equivalent to exploration with intrinsic bonus~\cite{schmidhuber1991curious, pathak2017curiosity, burda2018exploration}. We hope the framework of a dual system for exploration-exploitation tradeoff proposed in the current paper could inspire more systematic exploration methods currently unthought of.

 \vspace{-5pt}

\subsection*{Training}
\vspace{-5pt}
As we show in Section~\ref{sec:theory}, the learning processes for both agents converge to a stable solution. 
%
Note that since $\mathcal{A}^\explore=\mathcal{A}^\exploit$, {\fontfamily{cmss}\selectfont Exploiter} is trained off-policy using the data generated by the {\fontfamily{cmss}\selectfont Switcher} policy.
%
%
Depending on the nature of the action space of the environment
we can choose to implement SEREN with different baseline algorithms (e.g., value-based methods in discrete domains and policy gradient methods in continuous domains). Since we predominantly work with environments with continuous action spaces in Section~\ref{sec:experiments}, in Algorithm~\ref{algo:Opt_reward_shap_psuedo} we show the learning process of SEREN when we choose to implement {\fontfamily{cmss}\selectfont Switcher} based on Soft Actor-Critic~\cite{haarnoja2018soft}, a state-of-the-art off-policy policy gradient algorithms. Both {\fontfamily{cmss}\selectfont Exploiter} and {\fontfamily{cmss}\selectfont Switcher} are trained without exploration while {\fontfamily{cmss}\selectfont Switcher}'s interventions are determined according to the condition in Prop.~\ref{prop:switching_times}.




\begin{algorithm}[h!]\textbf{Algorithm 1 }
\begin{algorithmic}[1] 
\STATE Given reward objective function for {\fontfamily{cmss}\selectfont Switcher}, uncertainty objective function $\mathbf{L}(\cdot, \cdot)$, initialise Replay Buffer $\mathcal{B}$, \switcher~intervention cost $\beta$, 
		\FOR{$N_{episodes}$}
		    \STATE Reset state $s_0$
		    \FOR{$t=0,1,\ldots$}
    		    \STATE {\fontfamily{cmss}\selectfont Exploiter} sample $a^\exploit_{t}\sim\pi^\exploit(\cdot|s_{t})$ 
    		    \STATE {\fontfamily{cmss}\selectfont Explorer} action $a^\explore_{t} = \pi^\explore(s_t)$
    		      \STATE sample $g$ from $\mathfrak{g}(s_t)$
    		    \IF{$g =0$} 
         \STATE Apply $a^\exploit_{t}$ so $s_{t+1}\sim P(\cdot|a^\exploit_t,s_t),$
        		    \STATE Receive rewards $r^\exploit_{t} = R(s_{t},a^\exploit_{t})$ and $r^\explore_t = - \boldsymbol{L}(s_t,a^\exploit_t)$.
                    \STATE Store $(s_t, a^\exploit_{t}, s_{t+1}, r^\exploit_t, r^\explore_t)$ in $\mathcal{B}$
    		    \ELSE
        		    \STATE Apply $a^\explore_{t}$ so $s_{t+1}\sim P(\cdot|a^\explore_t,s_t)$
        		    \STATE Receive rewards $r^\exploit_t = R(s_{t},a^\explore_{t})$ and  $r^\explore_t = -\boldsymbol{L}(s_t,a^\explore_t) - \beta\cdot\boldsymbol{1}_{\mathcal{A}^\explore}(a^\explore_t)$ (Eq.~\ref{p3objective}).
                    \STATE Store $(s_t, a^\explore_t, s_{t+1}, r^\exploit_t, r^\explore_t)$ in $\mathcal{B}$
    		    \ENDIF
        	\ENDFOR
    	\STATE{\textbf{// Learn the individual policies}}
    	\STATE Sample a batch of $|B|$ transitions $B = \{(s_t, a_t, s_{t+1}, r_{t}, r^\explore_t)_{b}\}$ from $\mathcal{B}$
        \STATE Update {\fontfamily{cmss}\selectfont Switcher}'s policy using $B$ 
        \STATE Update {\fontfamily{cmss}\selectfont Exploiter}'s policy using $B$ 
        \ENDFOR
	\caption{\textbf{SE}lective \textbf{R}einforcement \textbf{E}xploration \textbf{N}etwork  (SEREN)}
	\label{algo:Opt_reward_shap_psuedo} 
\end{algorithmic}     

\end{algorithm}

%
%
%

\textbf{Other Learning Aspects}

    
    
    $\bullet$ The intervention criterion is determined by the `obstacle' switching condition \eqref{obstacle_condition} of Prop. 1. In order to execute interventions that closely adhere to the condition \eqref{obstacle_condition}, we use methods that ensure that the obstacle condition in Prop. 1. is accurate. For this, Q learning is a good candidate (and not critic methods with non-linear function approximator) for \switcher.
    
    $\bullet$ Our switch is according to the deterministic rule, Prop. 1.

\vspace{-5pt}
\section{Convergence \& Optimality of SEREN} \label{sec:theory}
\vspace{-5pt}
A key aspect of our framework is the presence of two RL agents that each adapt their play according to the other's behaviour. This produces two concurrent learning processes each designed to fulfill distinct objectives. At a stable point of the learning processes the {\fontfamily{cmss}\selectfont Switcher} minimises uncertainty about unexplored states while {\fontfamily{cmss}\selectfont Exploiter} maximises the environment reward.  However, introducing simultaneous learners occasions issues that generally prevent convergence to the stable point \cite{zinkevich2006cyclic}. 

We now show $\mathcal{G}$ admits a stable point and that our method converges to it. In particular, we show that the joint system converges in its value functions for each agent. Additionally, we show that SEREN induces a natural schedule in which as the environment is explored, {\fontfamily{cmss}\selectfont Switcher}'s interventions (to perform exploration) tend to $0$. We solve these challenges with the following scheme of results: 

\textbf{[A]}  Given any {\fontfamily{cmss}\selectfont Exploiter} policy, the {\fontfamily{cmss}\selectfont Switcher}'s learning process converges. 

\textbf{[B]} The switch activations performed by {\fontfamily{cmss}\selectfont Switcher} can be characterised by a `single obstacle condition' which can be evaluated online. Moreover, the number of switch activations  tends to $0$ as the system uncertainty decreases.

\textbf{[C]}  The system of two joint learners (SEREN) converges, moreover, SEREN converges to an approximate solution using function approximators for the critic.

We begin by stating a key result:

\begin{theorem}\label{convergence_theorem}
SEREN converges to a stable solution in the agents' value functions.
\end{theorem}

Theorem \ref{convergence_theorem} is established by proving a series of results; firstly that for a given {\fontfamily{cmss}\selectfont Exploiter} policy, {\fontfamily{cmss}\selectfont Switcher}'s learning process converges (to its optimal value function). Secondly, we show that the system of the two learners {\fontfamily{cmss}\selectfont Exploiter} and {\fontfamily{cmss}\selectfont Switcher} jointly converges to their optimal value functions.

Our first result proves that the {\fontfamily{cmss}\selectfont Switcher}'s optimal value function can be obtained as a limit point of a sequence of Bellman operations. We then prove that its convergence extends to the case with function approximators. To begin, first define a \textit{projection} $\Pi$ by: $
\Pi \Lambda:=\underset{\bar{\Lambda}\in\{\Phi r|r\in\mathbb{R}^p\}}{\arg\min}\left\|\bar{\Lambda}-\Lambda\right\|$ for any function $\Lambda$.

\begin{proposition}\label{convergence_switcher}
For a given {\fontfamily{cmss}\selectfont Exploiter} policy $\pi\in\Pi$, the {\fontfamily{cmss}\selectfont Switcher}'s learning process converges, moreover 
given a set of linearly independent basis functions $\Phi=\{\phi_1,\ldots,\phi_p\}$ where $\phi_{1\leq k\leq p}\in L_2$, the {\fontfamily{cmss}\selectfont Switcher}'s value function converges to a limit point $r^\star\in\mathbb{R}^p$ which is the unique solution to  $\Pi \mathfrak{F} (\Phi r^\star)=\Phi r^\star$ where $\mathfrak{F}$ is defined by:
    $\mathfrak{F}\Lambda:=R+\gamma P \max\{\mathcal{M}\Lambda,\Lambda\}$ where $\mathcal{M}$ is the {\fontfamily{cmss}\selectfont Switcher}'s intervention operator (c.f. \eqref{intervention_op}). Moreover, $r^\star$ satisfies: $
    \left\|\Phi r^\star - Q_2^\star\right\|\leq (1-\gamma^2)^{-1/2}\left\|\Pi Q_2^\star-Q_2^\star\right\|$.
\end{proposition}

Prop. \ref{convergence_switcher} establishes the convergence of the {\fontfamily{cmss}\selectfont Switcher}'s learning process with the use of a function approximator. The second statement bounds the proximity of the convergence point by the smallest approximation error that can be achieved given the choice of basis functions.

Having constructed a procedure to find the optimal {\fontfamily{cmss}\selectfont Exploiter} policy, our next result characterises the {\fontfamily{cmss}\selectfont Switcher} policy $\mathfrak{g}$ and the times that {\fontfamily{cmss}\selectfont Switcher} must perform an intervention. 
\begin{proposition}\label{prop:switching_times}

i) For any $s\in\mathcal{S}$, the {\fontfamily{cmss}\selectfont Switcher} intervention times are given by the following: 
\begin{smalleralign}
\tau_k=\inf\left\{\tau>\tau_{k-1}|\mathcal{M}^{\pi^\explore}v_2^{\pi^\exploit,\pi^\explore}= v_2^{\pi^\exploit,\Pi^\explore}\right\} \label{obstacle_condition}
\end{smalleralign}

ii) Denote by $ \mu_{l}(\mathfrak{g}) $ the number of switch activations performed by the {\fontfamily{cmss}\selectfont Switcher} when $\underset{(s,a)\in\mathcal{S}\times\mathcal{A}}{\max}L(s,a)=l$ under the  {\fontfamily{cmss}\selectfont Switcher} policy $\mathfrak{g}$, then $\underset{l\to 0}{\lim}\mu_{l}(\mathfrak{g})=0$. 
%
%
\vspace{-5pt}
\end{proposition} 
Part i) of Prop. \ref{prop:switching_times} characterises the distribution $\mathfrak{g}$. Moreover, given the function $V$, the times $\{\tau_k\}$ can be determined by evaluating if $\mathcal{M}V=V$ holds. Part ii) of Prop. \ref{prop:switching_times} establishes that the number of switches performed by \switcher~tends to $0$ as the system uncertainty is reduced through \switcher's exploration. This induces a natural exploration schedule based on the current system uncertainty.  

All proofs can be found in Appendix~\ref{sec:proofs_appendix}.
\vspace{-5pt}

\subsection*{Relation to Markov games.} 
Our framework involves a system of two agents each with their individual objectives. Settings of this kind are formalised by Markov games (MG), a framework for studying self-interested agents that simultaneously act over time  \cite{littman1994markov}. In the standard MG setup, the actions of \textit{both} agents influence both each agent's rewards and the system dynamics. Therefore, each agent $i\in\{1,2\}$ has its own reward function $R_i:\mathcal{S}\times(\times_{i=1}^2\mathcal{A}_i)\to\mathbb{R}$ and action set $\mathcal{A}_i$ and its goal is to maximise its \textit{own} expected returns. The system dynamics, now influenced by both agents, are described by a transition probability $P:\mathcal{S} \times(\times_{i=1}^2\mathcal{A}_i) \times \mathcal{S} \rightarrow [0, 1]$. Unlike classical MGs, in our MG, {\fontfamily{cmss}\selectfont Switcher} does not intervene at each state but is allowed to assume control of the system at certain states which it decides using impulse controls.  
\vspace{-5pt}

\section{Experiments} \label{sec:experiments}

We performed a series of experiments which demonstrate that SEREN's multi-player framework is able to improve the tradeoff between exploration and exploitation leading to marked improvement of the underling RL methods (further experimental details can be found in Appendix~\ref{sec: implementation_details}).

Specifically, we wish to address the following questions. \textbf{1.} Does SEREN learn to improve performance of an underlying base RL learner by more efficiently locating higher reward states under a) discrete b) continuous MDPs with different base learners (e.g., value-based and actor-critic)?
\textbf{2. }Does the non-stationary reward structure negatively impact the overall learning?
\textbf{3. }To what extent is SEREN agnostic with respect to the choice of the exploration bonus (Eq.~\ref{eq: ensemble-var}).


\subsection{MiniGrid Environments}

\begin{wrapfigure}{r}{0.5\textwidth}
\vspace{-10pt}
     \centering
     \begin{subfigure}[b]{0.18\textwidth}
         \centering
         \includegraphics[width=\textwidth]{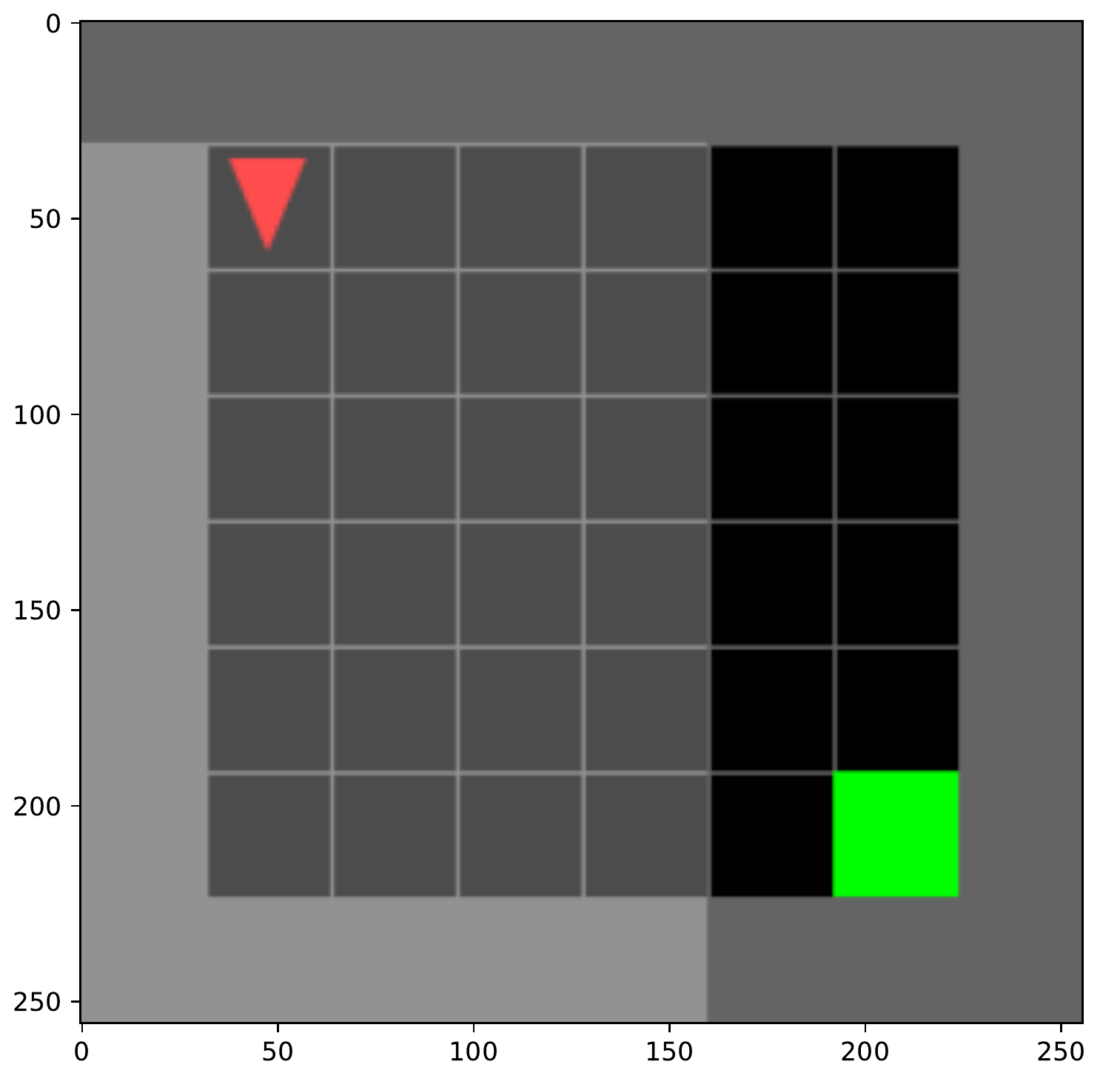}
         \caption{}
         \label{fig: minigrid_env}
     \end{subfigure}
     \hfill
     \begin{subfigure}[b]{0.29\textwidth}
         \centering
         \includegraphics[width=\textwidth]{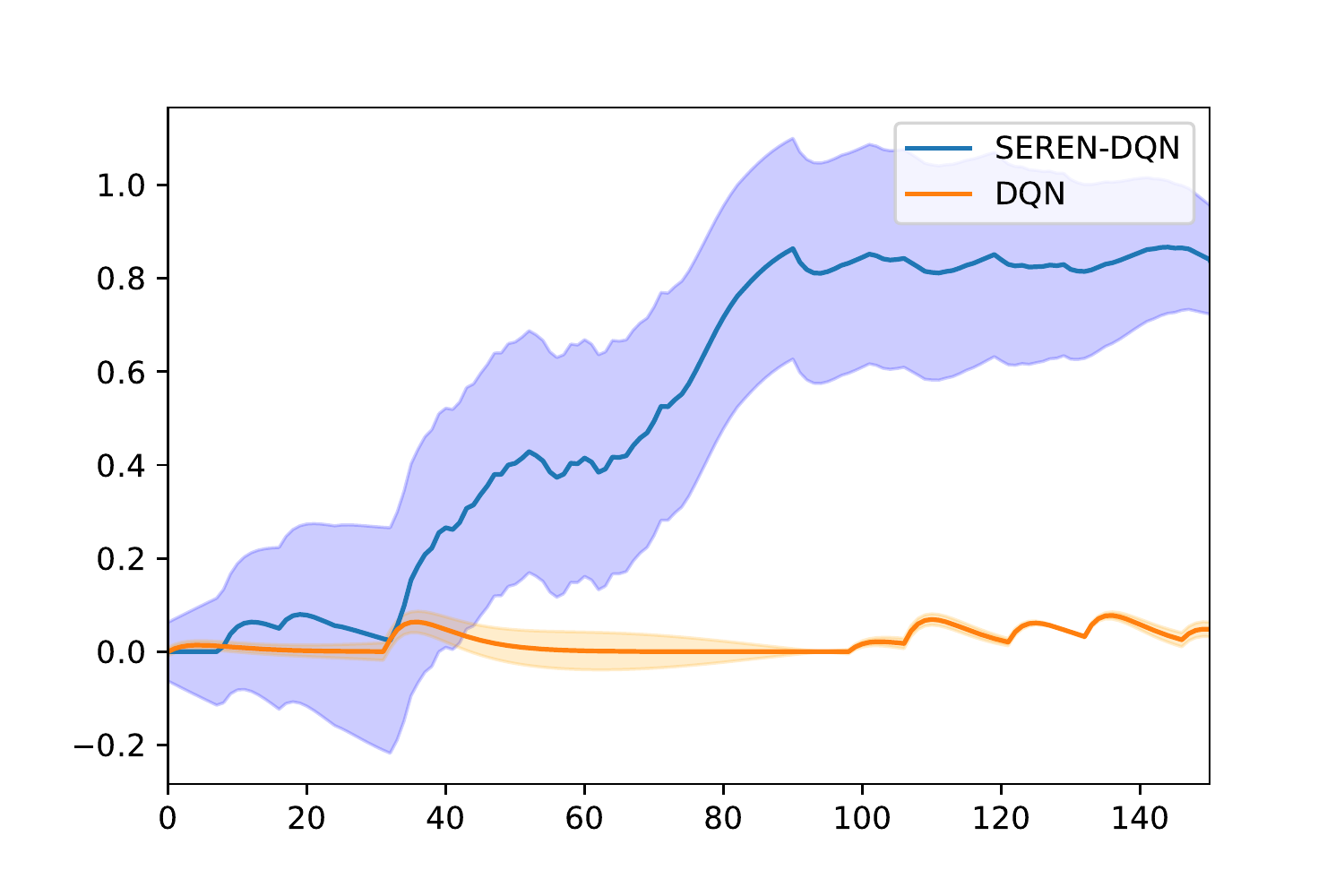}
         \caption{}
         \label{fig: DQN_comparison}
     \end{subfigure}
        \caption{DQN and SEREN-DQN in the MiniGrid-World~\cite{gym_minigrid}. (a) Graphical illustration of the ``8x8" minigrid-world environment, only transitions into goal states (green block) lead to non-zero rewards; (b) SEREN-DQN quickly learns the optimal policy to the goal state while the standard DQN has not learned good policy over $150$ episodes of training.}
        \label{fig: dqn_minigrid}
        \vspace{-10pt}
\end{wrapfigure}

We firstly demonstrate SEREN in combination with a standard DQN~\cite{mnih2013playing}. It is well known that DQN usually performs poorly in sparse-reward settings~\cite{osband2016deep, pathak2017curiosity}. To this end, we choose the MiniGrid environments~\cite{gym_minigrid},
where all transitions to non-goal states leads to zero reward. As we observe in Figure~\ref{fig: DQN_comparison}, SEREN-DQN quickly learns to consistently navigate towards the goal state within $100$ training episodes, whereas the standard DQN with $\epsilon$-greedy has failed to acquire a sensible policy over the $150$ episodes. Hence we conclude that SEREN can be readily plugged into DQNs to deal with sparse-reward and/or goal-directed tasks.

\vspace{-5pt}

\subsection{MuJoCo Environments}

\begin{figure}[h!]
    \centering
    \begin{subfigure}{.92\linewidth}
         \centering
         \includegraphics[width=\linewidth]{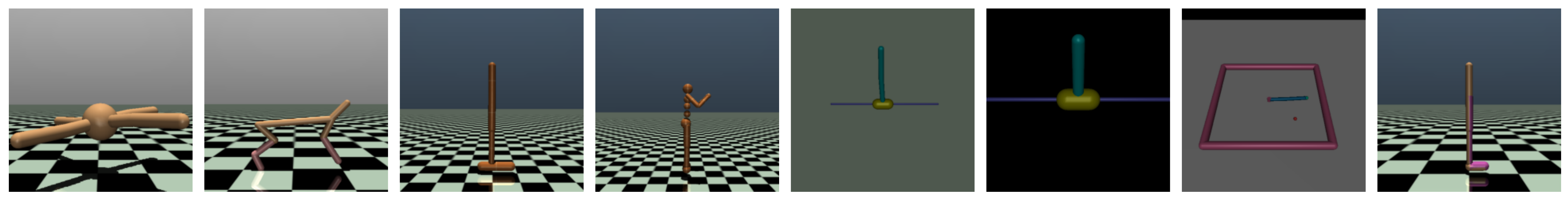}
         \caption{}
         \label{fig: mujoco_envs}
     \end{subfigure}
     \begin{subfigure}{.92\linewidth}
         \centering
    \includegraphics[width=\linewidth]{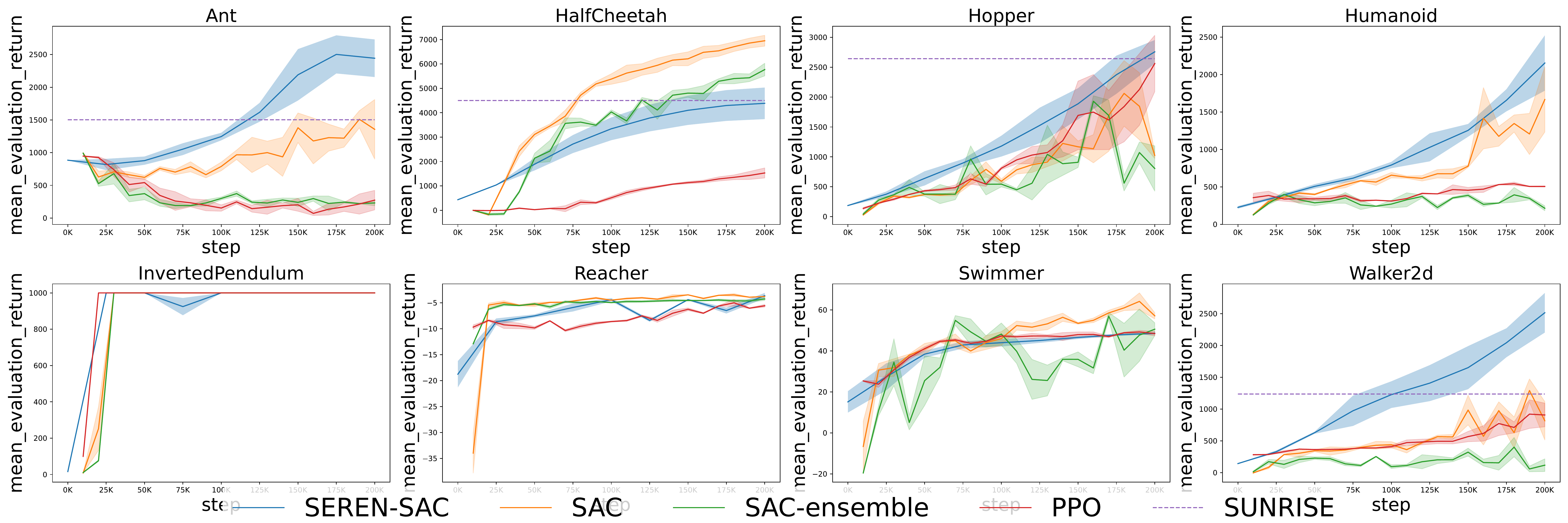}
         \caption{}
         \label{fig: sac_mujoco}
     \end{subfigure}
    \caption{Evaluation of SEREN with the baseline  SAC and TD3 algorithms on the MuJoCo tasks~\cite{todorov2012mujoco}. (a) Graphical demonstration of the selected MuJoCo environments; (b) Average evaluation returns (over $5$ random seeds) of SEREN-SAC over the $2\times10^5$ training steps.}
    \label{fig: main_exp}
    \vspace{-10pt}
\end{figure}

We next evaluate SEREN on the continuous control benchmarks from the MuJoCo suite \cite{todorov2012mujoco} (Figure~\ref{fig: mujoco_envs};~\citet{todorov2012mujoco}) to show that SEREN yields an exploration strategy that enables more efficient acquisition of the optimal policy. We choose to implement SEREN on the Soft Actor-Critic~\footnote{Our implementation is based on \textit{Stable-Baselines3} library~\cite{stable-baselines3}, \url{https://github.com/DLR-RM/stable-baselines3}} (SAC; \citet{haarnoja2018soft}). SAC is an off-policy policy gradient algorithm where the policy is trained under the maximum entropy principle, and it achieves state-of-the-art performance across a number of continuous control benchmarks.

From Figure~\ref{fig: sac_mujoco}, we observe that in $6$ out of the $8$ selected tasks, SEREN-SAC outperforms or is comparable with the baseline agents (SAC~\cite{haarnoja2018soft}, PPO~\cite{schulman2017Proximal}, SUNRISE~\cite{lee2021sunrise}) in terms of sample efficiency over the first $2\times10^5$ training steps. Specifically, we observe significant improvement over baseline SAC on $4$ tasks (Ant, Hopper, Humanoid, Walker2d). 
The gain may be attributed to the effective exploration by the \switcher, especially during the early phase of training, which facilitates the diversity of the off-policy replay buffer, hence enabling the identifying of better solutions. We show the comparison of asymptotic performance at $10^6$ training steps in Appendix~\ref{sec: additional_experiment}.

In order to empirically evaluate the utility of the impulse switching mechanism over intrinsic exploration, where the exploration bonus is combined with the extrinsic reward signal to guide exploration~\cite{schmidhuber1991curious, pathak2017curiosity}, we implement SAC-ensemble (green curves in Figure~\ref{fig: sac_mujoco}), that resembles the standard SAC with the critic function parameterised by an ensemble of neural networks, and the reward is augmented with the disagreement of the critic ensemble predictions (i.e., the same exploration objective as SEREN-SAC, Eq.~\ref{eq:q-ensemble-var}). We see that SEREN-SAC outperforms or is comparable with SAC-Ensemble on $7$ out of the $8$ tasks, hence demonstrating that including the impulse switching control enables more targeted exploration than naive combination of the extrinsic reward and the exploration bonus. 

We further demonstrate the flexibility of SEREN to accommodate different base learners in Appendix~\ref{sec: additional_experiment}, where we show that the combination of SEREN with another state-of-the-art off-policy RL algorithm, TD3~\cite{fujimoto2018addressing} also enables significant improvement in sample efficiency, illustrating the agnostic nature of SEREN with respect to the base learners.

\vspace{-5pt}

\subsection{Ablation Studies}

\begin{figure}[h!]
    \centering
    \begin{subfigure}{.48\linewidth}
         \centering
         \includegraphics[width=\linewidth]{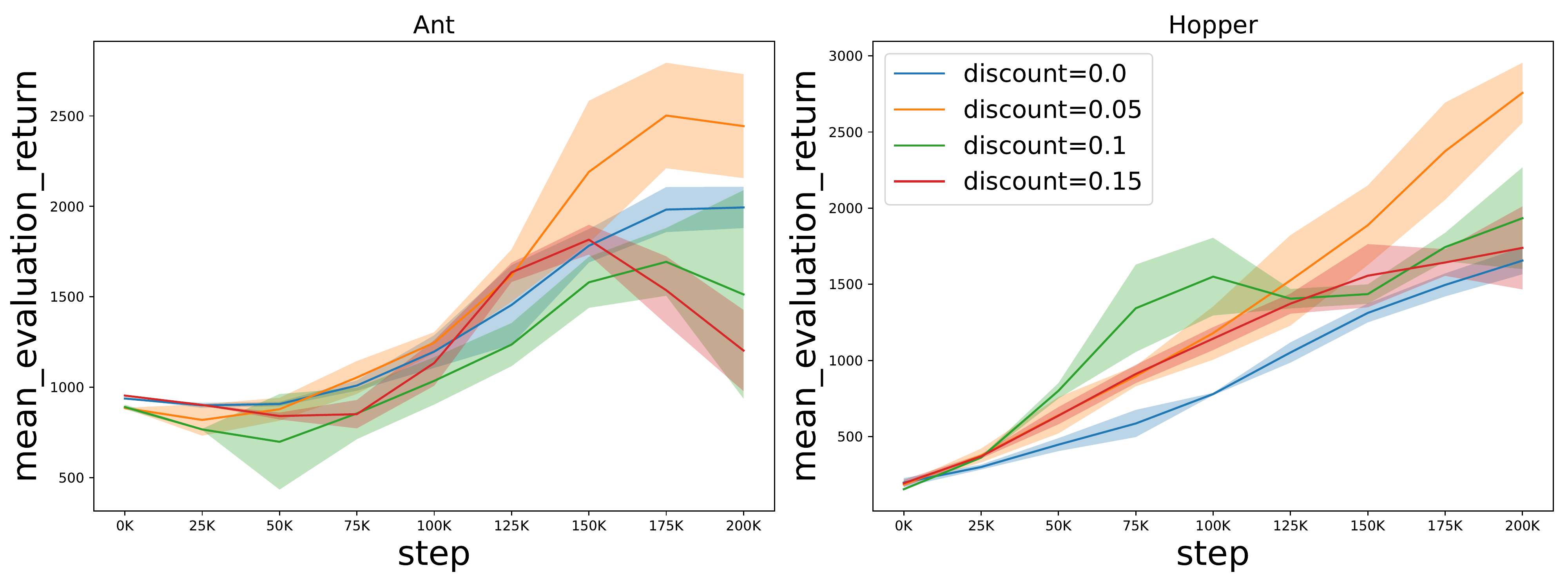}
         \vspace{-15pt}
         \caption{}
         \label{fig: sac_discount}
     \end{subfigure}
    \hfill
     \begin{subfigure}{.48\linewidth}
         \centering
    \includegraphics[width=\linewidth]{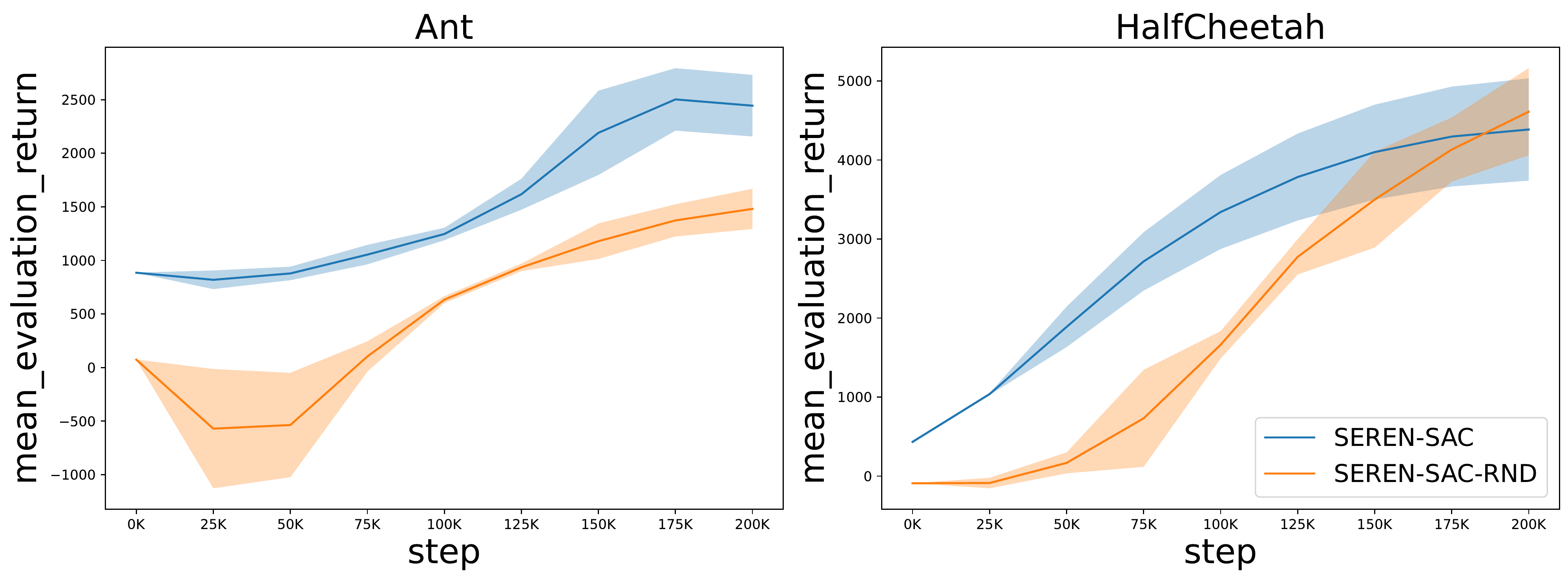}
    \vspace{-15pt}
         \caption{}
         \label{fig: sac_rnd}
     \end{subfigure}
     \vspace{-5pt}
    \caption{Ablations studies on SEREN-SAC with respect to the effect of (a) discounting factor of {\fontfamily{cmss}\selectfont Explorer} learning; (b) different exploration objective function.}
    \label{fig: sac_ablation}
\end{figure}

\textbf{Explorer Discount Factor and Non-Stationary Reward Structure of \switcher.}
As discussed in Sec.~\ref{sec: method}, the discounting factor for the {\fontfamily{cmss}\selectfont Explorer} needs to be set small so that the agent is myopic with respect to rewards in distant states to mitigate the effects of a non-stationarity reward. However, naively setting the discounting factor to $0$ would not yield good performance either, where the resulting agent takes exploratory actions only dependent on the epistemic uncertainty of the current state instead of a value estimate that guides targeted exploration.
Here we empirically justify our hypothesis by performing an ablation study on the effect of the value of the discounting factor for the learning of the {\fontfamily{cmss}\selectfont Explorer}. In Figure~\ref{fig: sac_discount}, we observe that by setting $\gamma_{2}=0.05$ (which we adopted in all experiments shown in Figure~\ref{fig: sac_mujoco}), there are noticeable performance improvement over other settings (including setting $\gamma_{2}=0.0$).
Hence the empirical evidence confirms our hypothesis that the non-stationary reward (epistemic uncertainty) structure of the {\fontfamily{cmss}\selectfont Explorer} learning can be ameliorated by setting the discounting factor appropriately small.
\vspace{-5pt}

\textbf{Different Exploration Objectives.} Different exploration objective leads to different behaviours. Here we wish to evaluate to what extent SEREN is agnostic to change the exploration objective. From Figure~\ref{fig: sac_rnd} we observe that $\text{SEREN-SAC}_{\text{RND}}$ is outperformed by standard SEREN-SAC with epistemic uncertainty bonus on both shown tasks, but still reaches comparable or better performance than the baselines (SAC and SUNRISE). Hence despite not being perfectly agnostic with respect to the exploration objective, the phenomenon that SEREN improves upon the base learner is not affected by the choice of the exploration bonus.

\vspace{-15pt}
\section{Discussion}
\vspace{-5pt}
\label{Section:Conclusion}
We introduced SEREN, an easy-to-plug \& play method that seeks to learn the optimal exploration-exploitation trade-off using an impulse control mechanism. SEREN can be readily combined with many existing value-based and policy gradient algorithms, such as DQN, SAC, TD3 (appendix~\ref{sec: additional_experiment}), etc. We formulate the problem of exploration-exploitation trade-off under a Markov game framework, where the {\fontfamily{cmss}\selectfont Exploiter} seeks to only maximise the cumulative return and the {\fontfamily{cmss}\selectfont Explorer} seeks to minimise the epistemic uncertainty of the {\fontfamily{cmss}\selectfont Exploiter}'s value estimate over the state space. We provide theoretical justification for the convergence of SEREN to the optimal achievable value estimates with linear function approximator. We demonstrate the utility of SEREN through extensive experimental studies on continuous control benchmarks. When implemented with state-of-the-art policy gradient algorithms (SAC), we show that the SEREN-augmented agents consistently yield improvement in terms of sample efficiency and asymptotic performance with respect to the baseline agents. We also showed that SEREN can be combined with value-based algorithms such as DQN, and yield significant improvement on sparse-reward environments, where standard DQNs usually fail to solve. 

Behaviourally animals tend to sacrifice short-term rewards to obtain information gain in uncertain environments~\cite{bromberg2009midbrain, gottlieb2013information}. \citet{blanchard2015orbitofrontal} demonstrated that the OFC neurons are the neural correlates encoding for both information value and primary value signals. 
Instead of integrating these variables to code subjective value, they found that OFC neurons tend to encode the two signals in an orthogonal manner. Hence despite being behaviourally similar, the dual system of independent value representation and learning in SEREN may provides a more biologically plausible framework for exploration-exploitation tradeoff than intrinsic exploration based on the combination of extrinsic primary reward structure and intrinsic estimate of information value.



\newpage 
\bibliography{main}

@article{chua2018deep,
  title={Deep reinforcement learning in a handful of trials using probabilistic dynamics models},
  author={Chua, Kurtland and Calandra, Roberto and McAllister, Rowan and Levine, Sergey},
  journal={arXiv preprint arXiv:1805.12114},
  year={2018}
}

@article{borkar1997stochastic,
  title={Stochastic approximation with two time scales},
  author={Borkar, Vivek S},
  journal={Systems \& Control Letters},
  volume={29},
  number={5},
  pages={291--294},
  year={1997},
  publisher={Elsevier}
}

@inproceedings{sekar2020planning,
  title={Planning to explore via self-supervised world models},
  author={Sekar, Ramanan and Rybkin, Oleh and Daniilidis, Kostas and Abbeel, Pieter and Hafner, Danijar and Pathak, Deepak},
  booktitle={International Conference on Machine Learning},
  pages={8583--8592},
  year={2020},
  organization={PMLR}
}

@article{haarnoja2018soft,
  title={Soft actor-critic algorithms and applications},
  author={Haarnoja, Tuomas and Zhou, Aurick and Hartikainen, Kristian and Tucker, George and Ha, Sehoon and Tan, Jie and Kumar, Vikash and Zhu, Henry and Gupta, Abhishek and Abbeel, Pieter and others},
  journal={arXiv preprint arXiv:1812.05905},
  year={2018}
}

@inproceedings{jin2020reward,
  title={Reward-free exploration for reinforcement learning},
  author={Jin, Chi and Krishnamurthy, Akshay and Simchowitz, Max and Yu, Tiancheng},
  booktitle={International Conference on Machine Learning},
  pages={4870--4879},
  year={2020},
  organization={PMLR}
}

@article{zhang2020task,
  title={Task-agnostic exploration in reinforcement learning},
  author={Zhang, Xuezhou and Singla, Adish and others},
  journal={arXiv preprint arXiv:2006.09497},
  year={2020}
}

@incollection{littman1994markov,
  title={{M}arkov games as a framework for multi-agent reinforcement learning},
  author={Littman, Michael L},
  booktitle={Machine learning proceedings 1994},
  pages={157--163},
  year={1994},
  publisher={Elsevier}
}

@article{zinkevich2006cyclic,
  title={Cyclic equilibria in {M}arkov games},
  author={Zinkevich, Martin and Greenwald, Amy and Littman, Michael},
  journal={Advances in Neural Information Processing Systems},
  volume={18},
  pages={1641},
  year={2006},
  publisher={MIT; 1998}
}

@article{cosso2013stochastic,
  title={Stochastic differential games involving impulse controls and double-obstacle quasi-variational inequalities},
  author={Cosso, Andrea},
  journal={SIAM Journal on Control and Optimization},
  volume={51},
  number={3},
  pages={2102--2131},
  year={2013},
  publisher={SIAM}
}

@article{mguni2019cutting,
  title={Cutting Your Losses: Learning Fault-Tolerant Control and Optimal Stopping under Adverse Risk},
  author={Mguni, David},
  journal={arXiv preprint arXiv:1902.05045},
  year={2019}
}

@article{tsitsiklis1999optimal,
  title={Optimal stopping of {M}arkov processes: Hilbert space theory, approximation algorithms, and an application to pricing high-dimensional financial derivatives},
  author={Tsitsiklis, John N and Van Roy, Benjamin},
  journal={IEEE Transactions on Automatic Control},
  volume={44},
  number={10},
  pages={1840--1851},
  year={1999},
  publisher={IEEE}
}

@book{sutton2018reinforcement,
  title={Reinforcement learning: An introduction},
  author={Sutton, Richard S and Barto, Andrew G},
  year={2018},
  publisher={MIT press}
}

@article{mguni2018viscosity,
  title={A Viscosity Approach to Stochastic Differential Games of Control and Stopping Involving Impulsive Control},
  author={Mguni, David},
  journal={arXiv preprint arXiv:1803.11432},
  year={2018}
}

@inproceedings{jaakkola1994convergence,
  title={Convergence of stochastic iterative dynamic programming algorithms},
  author={Jaakkola, Tommi and Jordan, Michael I and Singh, Satinder P},
  booktitle={Advances in neural information processing systems},
  pages={703--710},
  year={1994}
}

@article{deisenroth2011learning,
  title={Learning to control a low-cost manipulator using data-efficient reinforcement learning},
  author={Deisenroth, Marc Peter and Rasmussen, Carl Edward and Fox, Dieter},
  journal={Robotics: Science and Systems VII},
  pages={57--64},
  year={2011}
}

@book{benveniste2012adaptive,
  title={Adaptive algorithms and stochastic approximations},
  author={Benveniste, Albert and M{\'e}tivier, Michel and Priouret, Pierre},
  volume={22},
  year={2012},
  publisher={Springer Science \& Business Media}
}

@article{peng2017multiagent,
  title={Multiagent bidirectionally-coordinated nets: Emergence of human-level coordination in learning to play starcraft combat games},
  author={Peng, Peng and Wen, Ying and Yang, Yaodong and Yuan, Quan and Tang, Zhenkun and Long, Haitao and Wang, Jun},
  journal={arXiv preprint arXiv:1703.10069},
  year={2017}
}

@article{mnih2013playing,
  title={Playing atari with deep reinforcement learning},
  author={Mnih, Volodymyr and Kavukcuoglu, Koray and Silver, David and Graves, Alex and Antonoglou, Ioannis and Wierstra, Daan and Riedmiller, Martin},
  journal={arXiv preprint arXiv:1312.5602},
  year={2013}
}

@article{kingma2014adam,
  title={Adam: A method for stochastic optimization},
  author={Kingma, Diederik P and Ba, Jimmy},
  journal={arXiv preprint arXiv:1412.6980},
  year={2014}
}

@article{burda2018exploration,
  title={Exploration by random network distillation},
  author={Burda, Yuri and Edwards, Harrison and Storkey, Amos and Klimov, Oleg},
  journal={arXiv preprint arXiv:1810.12894},
  year={2018}
}

@article{schulman2017Proximal,
  author    = {John Schulman and
               Filip Wolski and
               Prafulla Dhariwal and
               Alec Radford and
               Oleg Klimov},
  title     = {Proximal Policy Optimization Algorithms},
  journal   = {CoRR},
  volume    = {abs/1707.06347},
  year      = {2017},
 
}

@inproceedings{pathak2017curiosity,
  title={Curiosity-driven Exploration by Self-supervised Prediction},
  author={Pathak, Deepak and Agrawal, Pulkit and Efros, Alexei A and Darrell, Trevor},
  booktitle={International Conference on Machine Learning (ICML)},
  pages={2778--2787},
  year={2017}
}

@article{mguni2018optimal,
  title={Optimal Selection of Transaction Costs in a Dynamic Principal-Agent Problem},
  author={Mguni, David},
  journal={arXiv preprint arXiv:1805.01062},
  year={2018}
}

@inproceedings{schmidhuber1991curious,
  title={Curious model-building control systems},
  author={Schmidhuber, J{\"u}rgen},
  booktitle={Proc. international joint conference on neural networks},
  pages={1458--1463},
  year={1991}
}

@article{schmidhuber1997shifting,
  title={Shifting inductive bias with success-story algorithm, adaptive Levin search, and incremental self-improvement},
  author={Schmidhuber, J{\"u}rgen and Zhao, Jieyu and Wiering, Marco},
  journal={Machine Learning},
  volume={28},
  number={1},
  pages={105--130},
  year={1997},
  publisher={Springer}
}

@inproceedings{ratzlaff2020implicit,
  title={Implicit generative modeling for efficient exploration},
  author={Ratzlaff, Neale and Bai, Qinxun and Fuxin, Li and Xu, Wei},
  booktitle={International Conference on Machine Learning},
  pages={7985--7995},
  year={2020},
  organization={PMLR}
}

@inproceedings{jiang2020generative,
  title={Generative Exploration and Exploitation},
  author={Jiang, Jiechuan and Lu, Zongqing},
  booktitle={Proceedings of the AAAI Conference on Artificial Intelligence},
  volume={34},
  number={04},
  pages={4337--4344},
  year={2020}
}

@article{auer2002using,
  title={Using confidence bounds for exploitation-exploration trade-offs},
  author={Auer, Peter},
  journal={Journal of Machine Learning Research},
  volume={3},
  number={Nov},
  pages={397--422},
  year={2002}
}

@article{osband2016deep,
  title={Deep exploration via bootstrapped DQN},
  author={Osband, Ian and Blundell, Charles and Pritzel, Alexander and Van Roy, Benjamin},
  journal={Advances in neural information processing systems},
  volume={29},
  pages={4026--4034},
  year={2016}
}

@article{lillicrap2015continuous,
  title={Continuous control with deep reinforcement learning},
  author={Lillicrap, Timothy P and Hunt, Jonathan J and Pritzel, Alexander and Heess, Nicolas and Erez, Tom and Tassa, Yuval and Silver, David and Wierstra, Daan},
  journal={arXiv preprint arXiv:1509.02971},
  year={2015}
}

@article{schwartenbeck2013exploration,
  title={Exploration, novelty, surprise, and free energy minimization},
  author={Schwartenbeck, Philipp and FitzGerald, Thomas and Dolan, Ray and Friston, Karl},
  journal={Frontiers in psychology},
  volume={4},
  pages={710},
  year={2013},
  publisher={Frontiers}
}

@inproceedings{todorov2012mujoco,
  title={Mujoco: A physics engine for model-based control},
  author={Todorov, Emanuel and Erez, Tom and Tassa, Yuval},
  booktitle={2012 IEEE/RSJ International Conference on Intelligent Robots and Systems},
  pages={5026--5033},
  year={2012},
  organization={IEEE}
}

@article{stable-baselines3,
  author  = {Antonin Raffin and Ashley Hill and Adam Gleave and Anssi Kanervisto and Maximilian Ernestus and Noah Dormann},
  title   = {Stable-Baselines3: Reliable Reinforcement Learning Implementations},
  journal = {Journal of Machine Learning Research},
  year    = {2021},
  volume  = {22},
  number  = {268},
  pages   = {1-8},
  url     = {http://jmlr.org/papers/v22/20-1364.html}
}

@inproceedings{fujimoto2018addressing,
  title={Addressing function approximation error in actor-critic methods},
  author={Fujimoto, Scott and Hoof, Herke and Meger, David},
  booktitle={International Conference on Machine Learning},
  pages={1587--1596},
  year={2018},
  organization={PMLR}
}

@inproceedings{janner2019trust,
  title={When to Trust Your Model: Model-Based Policy Optimization},
  author={Janner, Michael and Fu, Justin and Zhang, Marvin and Levine, Sergey},
  booktitle={Advances in Neural Information Processing Systems},
  volume={32},
  pages={12519--12530},
  year={2019}
}

@inproceedings{lee2021sunrise,
  title={Sunrise: A simple unified framework for ensemble learning in deep reinforcement learning},
  author={Lee, Kimin and Laskin, Michael and Srinivas, Aravind and Abbeel, Pieter},
  booktitle={International Conference on Machine Learning},
  pages={6131--6141},
  year={2021},
  organization={PMLR}
}

@article{Stadie2015IncentivizingEI,
  title={Incentivizing Exploration In Reinforcement Learning With Deep Predictive Models},
  author={Bradly C. Stadie and Sergey Levine and P. Abbeel},
  journal={ArXiv},
  year={2015},
  volume={abs/1507.00814}
}

@misc{gym_minigrid,
  author = {Chevalier-Boisvert, Maxime and Willems, Lucas and Pal, Suman},
  title = {Minimalistic Gridworld Environment for OpenAI Gym},
  year = {2018},
  publisher = {GitHub},
  journal = {GitHub repository},
  howpublished = {\url{https://github.com/maximecb/gym-minigrid}},
}

@article{gottlieb2013information,
  title={Information-seeking, curiosity, and attention: computational and neural mechanisms},
  author={Gottlieb, Jacqueline and Oudeyer, Pierre-Yves and Lopes, Manuel and Baranes, Adrien},
  journal={Trends in cognitive sciences},
  volume={17},
  number={11},
  pages={585--593},
  year={2013},
  publisher={Elsevier}
}

@article{blanchard2015orbitofrontal,
  title={Orbitofrontal cortex uses distinct codes for different choice attributes in decisions motivated by curiosity},
  author={Blanchard, Tommy C and Hayden, Benjamin Y and Bromberg-Martin, Ethan S},
  journal={Neuron},
  volume={85},
  number={3},
  pages={602--614},
  year={2015},
  publisher={Elsevier}
}

@article{o2021variational,
  title={Variational bayesian reinforcement learning with regret bounds},
  author={O'Donoghue, Brendan},
  journal={Advances in Neural Information Processing Systems},
  volume={34},
  year={2021}
}

@article{bromberg2009midbrain,
  title={Midbrain dopamine neurons signal preference for advance information about upcoming rewards},
  author={Bromberg-Martin, Ethan S and Hikosaka, Okihide},
  journal={Neuron},
  volume={63},
  number={1},
  pages={119--126},
  year={2009},
  publisher={Elsevier}
}
\bibliographystyle{icml2021}

\newpage



\section{Notation \& Assumptions}\label{sec:notation_appendix}

We assume that $\mathcal{S}$ is defined on a probability space $(\Omega,\mathcal{F},\mathbb{P})$ and any $s\in\mathcal{S}$ is measurable with respect
to the Borel $\sigma$-algebra associated with $\mathbb{R}^p$. We denote the $\sigma$-algebra of events generated by $\{s_t\}_{t\geq 0}$
by $\mathcal{F}_t\subset \mathcal{F}$. In what follows, we denote by $\left( \mathcal{V},\|\|\right)$ any finite normed vector space and by $\mathcal{H}$ the set of all measurable functions. 

The results of the paper are built under the following assumptions which are standard within RL and stochastic approximation methods:

\textbf{Assumption 1}
The stochastic process governing the system dynamics is ergodic, that is  the process is stationary and every invariant random variable of $\{s_t\}_{t\geq 0}$ is equal to a constant with probability $1$.

\textbf{Assumption 2}
The constituent functions of the players' objectives $R$ and $L$ are in $L_2$ (square-integrable functions).

\textbf{Assumption 3}
For any positive scalar $c$, there exists a scalar $\mu_c$ such that for all $s\in\mathcal{S}$ and for any $t\in\mathbb{N}$ we have: $
    \mathbb{E}\left[1+\|s_t\|^c|s_0=s\right]\leq \mu_c(1+\|s\|^c)$.

\textbf{Assumption 4}
There exists scalars $C_1$ and $c_1$ such that for any function $J$ satisfying $|J(s)|\leq C_2(1+\|s\|^{c_2})$ for some scalars $c_2$ and $C_2$ we have that: $
    \sum_{t=0}^\infty\left|\mathbb{E}\left[J(s_t)|s_0=s\right]-\mathbb{E}[J(s_0)]\right|\leq C_1C_2(1+\|s_t\|^{c_1c_2})$.

\textbf{Assumption 5}
There exists scalars $c$ and $C$ such that for any $s\in\mathcal{S}$ we have that: $
    |J(z,\cdot)|\leq C(1+\|z\|^c)$ for $J\in \{R,L\}$.
    
We also make the following finiteness assumption on set of switching control policies for {\fontfamily{cmss}\selectfont Switcher}:

\textbf{Assumption 6}
For any policy $\mathfrak{g}_c$, the total number of interventions is given by $K<\infty$.

In what follows, we denote by $\boldsymbol{\mathcal{A}}:=\mathcal{A}^\exploit\times\mathcal{A}^\explore$ and by $\boldsymbol{\Pi}:=\Pi^\exploit\times\Pi^\explore\times G$ where $G$ is the policy set of the \switcher's policy $\mathfrak{g}$.

\section{Proof of Technical Results}\label{sec:proofs_appendix}

We begin the analysis with some preliminary lemmata and definitions which are useful for proving the main results.

\begin{definition}{A.1}
An operator $T: \mathcal{V}\to \mathcal{V}$ is said to be a \textbf{contraction} w.r.t a norm $\|\cdot\|$ if there exists a constant $c\in[0,1[$ such that for any $V_1,V_2\in  \mathcal{V}$ we have that:
\begin{align}
    \|TV_1-TV_2\|\leq c\|V_1-V_2\|.
\end{align}
\end{definition}

\begin{definition}{A.2}
An operator $T: \mathcal{V}\to  \mathcal{V}$ is \textbf{non-expansive} if $\forall V_1,V_2\in  \mathcal{V}$ we have:
\begin{align}
    \|TV_1-TV_2\|\leq \|V_1-V_2\|.
\end{align}
\end{definition}

\begin{lemma} \label{max_lemma}
For any
$f: \mathcal{V}\to\mathbb{R}: \mathcal{V}\to\mathbb{R}$, we have that:
\begin{align}
\left\|\underset{a\in \mathcal{V}}{\max}\:f(a)-\underset{a\in \mathcal{V}}{\max}\: g(a)\right\| \leq \underset{a\in \mathcal{V}}{\max}\: \left\|f(a)-g(a)\right\|.    \label{lemma_1_basic_max_ineq}
\end{align}
\end{lemma}
\begin{proof}
We restate the proof given in \cite{mguni2019cutting}:
\begin{align}
f(a)&\leq \left\|f(a)-g(a)\right\|+g(a)\label{max_inequality_proof_start}
\\\implies
\underset{a\in \mathcal{V}}{\max}f(a)&\leq \underset{a\in \mathcal{V}}{\max}\{\left\|f(a)-g(a)\right\|+g(a)\}
\leq \underset{a\in \mathcal{V}}{\max}\left\|f(a)-g(a)\right\|+\underset{a\in \mathcal{V}}{\max}\;g(a). \label{max_inequality}
\end{align}
Deducting $\underset{a\in \mathcal{V}}{\max}\;g(a)$ from both sides of (\ref{max_inequality}) yields:
\begin{align}
    \underset{a\in \mathcal{V}}{\max}f(a)-\underset{a\in \mathcal{V}}{\max}g(a)\leq \underset{a\in \mathcal{V}}{\max}\left\|f(a)-g(a)\right\|.\label{max_inequality_result_last}
\end{align}
After reversing the roles of $f$ and $g$ and redoing steps (\ref{max_inequality_proof_start}) - (\ref{max_inequality}), we deduce the desired result since the RHS of (\ref{max_inequality_result_last}) is unchanged.
\end{proof}

\begin{lemma}{A.4}\label{non_expansive_P}
The probability transition kernel $P$ is non-expansive, that is:
\begin{align}
    \|PV_1-PV_2\|\leq \|V_1-V_2\|.
\end{align}
\end{lemma} 
\begin{proof}
The result is well-known e.g. \cite{tsitsiklis1999optimal}. We give a proof using the Tonelli-Fubini theorem and the iterated law of expectations, we have that:
\begin{align*}
&\|PJ\|^2=\mathbb{E}\left[(PJ)^2[s_0]\right]=\mathbb{E}\left(\left[\mathbb{E}\left[J[s_1]|s_0\right]\right]^2\right]
\leq \mathbb{E}\left[\mathbb{E}\left[J^2[s_1]|s_0\right]\right] 
= \mathbb{E}\left[J^2[s_1]\right]=\|J\|^2,
\end{align*}
where we have used Jensen's inequality to generate the inequality. This completes the proof.
\end{proof}

\section*{Proof of Theorem \ref{convergence_theorem}}
We begin by proving the following result:
\begin{proposition}\label{theorem:q_learning}
Define by $\cR^\explore(s_t,a^\exploit_t,a^\explore_t):=R^\explore\left(s_t,a^\exploit_t,a^\explore_t\right)  - \beta\cdot\boldsymbol{1}_{\mathcal{A}^\explore}(a^\explore_t)$ and consider the following Q learning variant:
\begin{align*}
    &Q_{2,t+1}(s_t,a^\exploit_t,a^\explore_t)=Q_{2,t}(s_t,a^\exploit_t,a^\explore_t)
\\&+\alpha_t(s_t,a^\explore_t)\left[\max\left\{\mathcal{M}^{\pi,\mathfrak{g}}Q_{2,t}(s_t,a^\exploit_t,a^\explore_t), \cR(s_t,a^\exploit_t,a^\explore_t)+\gamma\underset{a'\in\mathcal{A}}{\max}\;Q_{2,t}(s_{t+1},a^\exploit_t,a'^\explore)\right\}-Q_{2,t}(s_t,a^\exploit_t,a^\explore_t)\right],
\end{align*}
then for a fixed the {\fontfamily{cmss}\selectfont Exploiter} policy $\pi^\exploit$ and for a fixed $L$, $Q_{2,t}(s)$ converges to $Q_2^\star$ with probability $1$, where $s_t,s_{t+1}\in\cS$ and $a^\exploit_t\sim\pi^\exploit(\cdot|s_t)$ is {\fontfamily{cmss}\selectfont Exploiter}'s action.
\end{proposition}

\begin{proof}

We begin by defining some objects which are central to the analysis. 
For any $\pi^\exploit\in\Pi^\exploit$ and $\pi^\explore\in\Pi^\explore$, given the value function $v_2^{\pi^\exploit,\pi^\explore}:\mathcal{S}\to\mathbb{R}$, we define the \textit{intervention operator} $\mathcal{M}^{\pi^\exploit,\pi^\explore}$ by 
\begin{align}
\mathcal{M}^{\pi^\exploit,\pi^\explore}v^{\pi^\exploit,(\pi^\explore,\mathfrak{g})}_2(s_{\tau_k}):= R\left(s_{\tau_k},a^\explore_{\tau_k}\right)  - \beta\boldsymbol{1}_{\mathcal{A}^\explore}(a^\explore_{\tau_k})+\gamma\sum_{s'\in\mathcal{S}}P(s';a^\explore_{\tau_k},s_{\tau_k})v^{\pi^\exploit,(\pi^\explore,\mathfrak{g})}_2(s') \label{intervention_op}
\end{align}
for any $s_{\tau_k}\in\mathcal{S}$ and $\forall \tau_k$ where  $a^\explore_{\tau_k}\sim  \pi^\explore(\cdot|s_{\tau_k})$. 

Next, let us define the Bellman operator $T$ of acting on the value function $v_2^{\pi^\exploit,\pi^\explore}:\mathcal{S}\to\mathbb{R}$ by
\begin{align}
T \Lambda(s_{\tau_k}):=\max\left\{\mathcal{M}^{\boldsymbol{\pi}}\Lambda(s_{\tau_k}),\left[ R^\explore(s_{\tau_k},\boldsymbol{a})+\gamma\underset{\boldsymbol{a}\in\boldsymbol{\mathcal{A}}}{\max}\;\sum_{s'\in\mathcal{S}}P(s';\boldsymbol{a},s_{\tau_k})\Lambda(s')\right]\right\}\label{bellman_proof_start}
\end{align}

Our first result proves that the operator  $T$ is a contraction operator. First let us recall that the \textit{switching time} $\tau_k$ is defined recursively $\tau_k=\inf\{t>\tau_{k-1}|s_t\in A,\tau_k\in\mathcal{F}_t\}$ where $A=\{s\in \mathcal{S}|\mathfrak{g}(s_t)=1\}$.
To this end, we show that the following bounds holds:
\begin{lemma}\label{lemma:bellman_contraction}
The Bellman operator $T$ is a contraction, that is the following bound holds:
\begin{align*}
&\left\|T\psi-T\psi'\right\|\leq \gamma\left\|\psi-\psi'\right\|.
\end{align*}
\end{lemma}

In what follows and for the remainder of the script, we employ the following shorthands:
\begin{align*}
&\mathcal{P}^{\boldsymbol{a}}_{ss'}=:\sum_{s'\in\mathcal{S}}P(s';\boldsymbol{a},s), \quad\mathcal{P}^{\boldsymbol{\pi}}_{ss'}=:\sum_{\boldsymbol{a}\in\boldsymbol{\mathcal{A}}}\boldsymbol{\pi}(\boldsymbol{a}|s)\mathcal{P}^{\boldsymbol{a}}_{ss'}
\end{align*}

To prove that $T$ is a contraction, we consider the three cases produced by \eqref{bellman_proof_start}, that is to say we prove the following statements:

i) $\qquad\qquad
\left| R^\explore(s_{\tau_k},\boldsymbol{a})+\gamma\underset{\boldsymbol{a}\in\boldsymbol{\mathcal{A}}}{\max}\;\mathcal{P}^{\boldsymbol{a}}_{s's_t}\psi(s',\cdot)-\left( R^\explore(s_{\tau_k},\boldsymbol{a})+\gamma\underset{\boldsymbol{a}\in\boldsymbol{\mathcal{A}}}{\max}\;\mathcal{P}^{\boldsymbol{a}}_{s's_t}\psi'(s',\cdot)\right)\right|\leq \gamma\left\|\psi-\psi'\right\|$

ii) $\qquad\qquad
\left\|\mathcal{M}^{\boldsymbol{\pi}}\psi-\mathcal{M}^{\boldsymbol{\pi}}\psi'\right\|\leq    \gamma\left\|\psi-\psi'\right\|,\qquad \qquad$
  (and hence $\mathcal{M}$ is a contraction).

iii) $\qquad\qquad
    \left\|\mathcal{M}^{\boldsymbol{\pi}}\psi-\left[ R^\explore(\cdot,\boldsymbol{a})+\gamma\underset{\boldsymbol{a}\in\boldsymbol{\mathcal{A}}}{\max}\;\mathcal{P}^{\boldsymbol{a}}\psi'\right]\right\|\leq \gamma\left\|\psi-\psi'\right\|.
$

We begin by proving i).

Indeed, for any $a^\exploit\in\mathcal{A}^\exploit$ and $\forall s_t\in\mathcal{S}, \forall s'\in\mathcal{S}$ we have that 
\begin{align*}
&\left| R^\explore(s_{\tau_k},\boldsymbol{a})+\gamma\mathcal{P}^\pi_{s's_t}\psi(s',\cdot)-\left[ R^\explore(s_{\tau_k},\boldsymbol{a})+\gamma\underset{\boldsymbol{a}\in\boldsymbol{\mathcal{A}}}{\max}\;\;\mathcal{P}^{\boldsymbol{a}}_{s's_t}\psi'(s',\cdot)\right]\right|
\\\leq &\underset{\boldsymbol{a}\in\boldsymbol{\mathcal{A}}}{\max}\;\left|\gamma\mathcal{P}^{\boldsymbol{a}}_{s's_t}\psi(s',\cdot)-\gamma\mathcal{P}^{\boldsymbol{a}}_{s's_t}\psi'(s',\cdot)\right|
\\\leq &\gamma\left\|P\psi-P\psi'\right\|
\\\leq &\gamma\left\|\psi-\psi'\right\|,
\end{align*}
again using the fact that $P$ is non-expansive and Lemma \ref{max_lemma}.

We now prove ii).

For any $\tau\in\mathcal{F}$, define by $\tau'=\inf\{t>\tau|s_t\in A,\tau\in\mathcal{F}_t\}$. Now using the definition of $\mathcal{M}$ we have that for any $s_\tau\in\mathcal{S}$
\begin{align*}
&\left|(\mathcal{M}^{\boldsymbol{\pi}}\psi-\mathcal{M}^{\boldsymbol{\pi}}\psi')(s_{\tau})\right|
\\&\leq \underset{\boldsymbol{a}_\tau,\in \boldsymbol{\mathcal{A}}}{\max}    \Bigg|R^\explore(s_\tau,\boldsymbol{a}_\tau)-\beta\boldsymbol{1}_{\mathcal{A}^\explore}(a^\explore_t)+\gamma\mathcal{P}^{\boldsymbol{\pi}}_{s's_\tau}\mathcal{P}^{\boldsymbol{a}}\psi(s_{\tau})
-\left(R^\explore(s_\tau,\boldsymbol{a}_\tau)-\beta\boldsymbol{1}_{\mathcal{A}^\explore}(a^\explore_t)+\gamma\mathcal{P}^{\boldsymbol{\pi}}_{s's_\tau}\mathcal{P}^{\boldsymbol{a}}\psi'(s_{\tau})\right)\Bigg| 
\\&= \gamma\left|\mathcal{P}^{\boldsymbol{\pi}}_{s's_\tau}\mathcal{P}^{\boldsymbol{a}}\psi(s_{\tau})-\mathcal{P}^{\boldsymbol{\pi}}_{s's_\tau}\mathcal{P}^{\boldsymbol{a}}\psi'(s_{\tau})\right| 
\\&\leq \gamma\left\|P\psi-P\psi'\right\|
\\&\leq \gamma\left\|\psi-\psi'\right\|,
\end{align*}
using the fact that $P$ is non-expansive. The result can then be deduced easily by applying max on both sides.

We now prove iii). We split the proof of the statement into two cases:

\textbf{Case 1:} 
\begin{align}\mathcal{M}^{\boldsymbol{\pi}}\psi(s_{\tau})-\left(R^\explore(s_\tau,\boldsymbol{a}_\tau)+\gamma\underset{\boldsymbol{a}\in\boldsymbol{\mathcal{A}}}{\max}\;\mathcal{P}^{\boldsymbol{a}}_{s's_\tau}\psi'(s')\right)<0.
\end{align}

We now observe the following:
\begin{align*}
&\mathcal{M}^{\boldsymbol{\pi}}\psi(s_{\tau})-R^\explore(s_\tau,\boldsymbol{a}_\tau)+\gamma\underset{\boldsymbol{a}\in\boldsymbol{\mathcal{A}}}{\max}\;\mathcal{P}^{\boldsymbol{a}}_{s's_\tau}\psi'(s')
\\&\leq\max\left\{R^\explore(s_\tau,\boldsymbol{a}_\tau)+\gamma\mathcal{P}^{\boldsymbol{\pi}}_{s's_\tau}\mathcal{P}^{\boldsymbol{a}}\psi(s'),\mathcal{M}^{\boldsymbol{\pi}}\psi(s_{\tau})\right\}-R^\explore(s_\tau,\boldsymbol{a}_\tau)+\gamma\underset{\boldsymbol{a}\in\boldsymbol{\mathcal{A}}}{\max}\;\mathcal{P}^{\boldsymbol{a}}_{s's_\tau}\psi'(s')
\\&\leq \Bigg|\max\left\{R^\explore(s_\tau,\boldsymbol{a}_\tau)+\gamma\mathcal{P}^{\boldsymbol{\pi}}_{s's_\tau}\mathcal{P}^{\boldsymbol{a}}\psi(s'),\mathcal{M}^{\boldsymbol{\pi}}\psi(s_{\tau})\right\}-\max\left\{R^\explore(s_\tau,\boldsymbol{a}_\tau)+\gamma\underset{\boldsymbol{a}\in\boldsymbol{\mathcal{A}}}{\max}\;\mathcal{P}^{\boldsymbol{a}}_{s's_\tau}\psi'(s'),\mathcal{M}^{\boldsymbol{\pi}}\psi(s_{\tau})\right\}
\\&+\max\left\{R^\explore(s_\tau,\boldsymbol{a}_\tau)+\gamma\underset{\boldsymbol{a}\in\boldsymbol{\mathcal{A}}}{\max}\;\mathcal{P}^{\boldsymbol{a}}_{s's_\tau}\psi'(s'),\mathcal{M}^{\boldsymbol{\pi}}\psi(s_{\tau})\right\}-R^\explore(s_\tau,\boldsymbol{a}_\tau)+\gamma\underset{\boldsymbol{a}\in\boldsymbol{\mathcal{A}}}{\max}\;\mathcal{P}^{\boldsymbol{a}}_{s's_\tau}\psi'(s')\Bigg|
\\&\leq \Bigg|\max\left\{R^\explore(s_\tau,\boldsymbol{a}_\tau)+\gamma\underset{\boldsymbol{a}\in\boldsymbol{\mathcal{A}}}{\max}\;\mathcal{P}^{\boldsymbol{a}}_{s's_\tau}\psi(s'),\mathcal{M}^{\boldsymbol{\pi}}\psi(s_{\tau})\right\}-\max\left\{R^\explore(s_\tau,\boldsymbol{a}_\tau)+\gamma\underset{\boldsymbol{a}\in\boldsymbol{\mathcal{A}}}{\max}\;\mathcal{P}^{\boldsymbol{a}}_{s's_\tau}\psi'(s'),\mathcal{M}^{\boldsymbol{\pi}}\psi(s_{\tau})\right\}\Bigg|
\\&\qquad+\Bigg|\max\left\{R^\explore(s_\tau,\boldsymbol{a}_\tau)+\gamma\underset{\boldsymbol{a}\in\boldsymbol{\mathcal{A}}}{\max}\;\mathcal{P}^{\boldsymbol{a}}_{s's_\tau}\psi'(s'),\mathcal{M}^{\boldsymbol{\pi}}\psi(s_{\tau})\right\}-R^\explore(s_\tau,\boldsymbol{a}_\tau)+\gamma\underset{\boldsymbol{a}\in\boldsymbol{\mathcal{A}}}{\max}\;\mathcal{P}^{\boldsymbol{a}}_{s's_\tau}\psi'(s')\Bigg|
\\&\leq \gamma\underset{a\in\mathcal{A}}{\max}\;\left|\mathcal{P}^{\boldsymbol{\pi}}_{s's_\tau}\mathcal{P}^{\boldsymbol{a}}\psi(s')-\mathcal{P}^{\boldsymbol{\pi}}_{s's_\tau}\mathcal{P}^{\boldsymbol{a}}\psi'(s')\right|
\\&\qquad+\left|\max\left\{0,\mathcal{M}^{\boldsymbol{\pi}}\psi(s_{\tau})-\left(R^\explore(s_\tau,\boldsymbol{a}_\tau)+\gamma\underset{\boldsymbol{a}\in\boldsymbol{\mathcal{A}}}{\max}\;\mathcal{P}^{\boldsymbol{a}}_{s's_\tau}\psi'(s')\right)\right\}\right|
\\&\leq \gamma\left\|P\psi-P\psi'\right\|
\\&\leq \gamma\|\psi-\psi'\|,
\end{align*}
where we have used the fact that for any scalars $a,b,c$ we have that $
    \left|\max\{a,b\}-\max\{b,c\}\right|\leq \left|a-c\right|$ and the non-expansiveness of $P$.

\textbf{Case 2: }
\begin{align*}\mathcal{M}^{\boldsymbol{\pi}}\psi(s_{\tau})-\left(R^\explore(s_\tau,\boldsymbol{a}_\tau)+\gamma\underset{\boldsymbol{a}\in\boldsymbol{\mathcal{A}}}{\max}\;\mathcal{P}^{\boldsymbol{a}}_{s's_\tau}\psi'(s')\right)\geq 0.
\end{align*}

For this case, we have that
\begin{align*}
&\mathcal{M}^{\boldsymbol{\pi}}\psi(s_{\tau})-\left(R^\explore(s_\tau,\boldsymbol{a}_\tau)+\gamma\underset{\boldsymbol{a}\in\boldsymbol{\mathcal{A}}}{\max}\;\mathcal{P}^{\boldsymbol{a}}_{s's_\tau}\psi'(s')\right)
\\&\leq \mathcal{M}^{\boldsymbol{\pi}}\psi(s_{\tau})-\left(R^\explore(s_\tau,\boldsymbol{a}_\tau)+\gamma\underset{\boldsymbol{a}\in\boldsymbol{\mathcal{A}}}{\max}\;\mathcal{P}^{\boldsymbol{a}}_{s's_\tau}\psi'(s')\right)+\beta\boldsymbol{1}_{\mathcal{A}^\explore}(a^\explore_t)
\\&\leq R^\explore(s_\tau,\boldsymbol{a}_\tau)-\beta\boldsymbol{1}_{\mathcal{A}^\explore}(a^\explore_t)+\gamma\mathcal{P}^{\boldsymbol{\pi}}_{s's_\tau}\mathcal{P}^{\boldsymbol{a}}\psi(s')
\\&\qquad\qquad\qquad\qquad\quad-\left(R^\explore(s_\tau,\boldsymbol{a}_\tau)-\beta\boldsymbol{1}_{\mathcal{A}^\explore}(a^\explore_t)+\gamma\underset{\boldsymbol{a}\in\boldsymbol{\mathcal{A}}}{\max}\;\mathcal{P}^{\boldsymbol{a}}_{s's_\tau}\psi'(s')\right)
\\&\leq \gamma\underset{\boldsymbol{a}\in\boldsymbol{\mathcal{A}}}{\max}\;\left|\mathcal{P}^{\boldsymbol{\pi}}_{s's_\tau}\mathcal{P}^{\boldsymbol{a}}\left(\psi(s')-\psi'(s')\right)\right|
\\&\leq \gamma\left|\psi(s')-\psi'(s')\right|
\\&\leq \gamma\left\|\psi-\psi'\right\|,
\end{align*}
again using the fact that $P$ is non-expansive. Hence we have succeeded in showing that for any $\Lambda\in L_2$ we have that
\begin{align}
    \left\|\mathcal{M}^{\boldsymbol{\pi}}\Lambda-\underset{\boldsymbol{a}\in\boldsymbol{\mathcal{A}}}{\max}\;\left[ \psi(\cdot,a)+\gamma\mathcal{P}^{\boldsymbol{a}}\Lambda'\right]\right\|\leq \gamma\left\|\Lambda-\Lambda'\right\|.\label{off_M_bound_gen}
\end{align}
Gathering the results of the three cases gives the desired result.

We now make use of the following result:
\begin{theorem}[Theorem 1, pg 4 in \cite{jaakkola1994convergence}]
Let $\Xi_t(s)$ be a random process that takes values in $\mathbb{R}^n$ and given by the following:
\begin{align}
    \Xi_{2,t+1}(s)=\left(1-\alpha_t(s)\right)\Xi_{2,t}(s)\alpha_t(s)L_t(s),
\end{align}
then $\Xi_t(s)$ converges to $0$ with probability $1$ under the following conditions:
\begin{itemize}
\item[i)] $0\leq \alpha_t\leq 1, \sum_t\alpha_t=\infty$ and $\sum_t\alpha_t<\infty$
\item[ii)] $\|\mathbb{E}[L_t|\mathcal{F}_t]\|\leq \gamma \|\Xi_t\|$, with $\gamma <1$;
\item[iii)] ${\rm Var}\left[L_t|\mathcal{F}_t\right]\leq c(1+\|\Xi_t\|^2)$ for some $c>0$.
\end{itemize}
\end{theorem}
\begin{proof}
To prove the convergence in Theorem \ref{convergence_theorem}, we show (i) - (iii) hold. Condition (i) holds by choice of learning rate. It therefore remains to prove (ii) - (iii). We first prove (ii). For this, we consider our variant of the Q-learning update rule:
\begin{align*}
Q_{2,t+1}(s_t,\boldsymbol{a}_t)=Q_{2,t}&(s_t,\boldsymbol{a}_t)+\alpha_t(s_t,\boldsymbol{a}_t)\left[\max\left\{\mathcal{M}^{\boldsymbol{\pi}}Q_{2,t}(s_{\tau_k},\boldsymbol{a}), R^\explore(s_{\tau_k},\boldsymbol{a})+\gamma\underset{a'\in\mathcal{A}}{\max}\;Q_{2,t}(s',\boldsymbol{a'})\right\}-Q_{2,t}(s_t,\boldsymbol{a}_t)\right].
\end{align*}
After subtracting $Q_2^\star(s_t,\boldsymbol{a}_t)$ from both sides and some manipulation we obtain that:
\begin{align*}
&\Xi_{2,t+1}(s_t,\boldsymbol{a}_t)
\\&=(1-\alpha_t(s_t,\boldsymbol{a}_t))\Xi_{2,t}(s_t,\boldsymbol{a}_t)
\\&\qquad\qquad\qquad\qquad\;\;+\alpha_t(s_t,\boldsymbol{a}_t)\left[\max\left\{\mathcal{M}^{\boldsymbol{\pi}}Q_{2,t}(s_{\tau_k},\boldsymbol{a}), R^\explore(s_{\tau_k},\boldsymbol{a})+\gamma\underset{a'\in\mathcal{A}}{\max}\;Q_{2,t}(s',\boldsymbol{a'})\right\}-Q_{2,t}^\star(s_t,\boldsymbol{a}_t)\right],  \end{align*}
where $\Xi_{2,t}(s_t,\boldsymbol{a}_t):=Q_{2,t}(s_t,\boldsymbol{a}_t)-Q_{2,t}^\star(s_t,\boldsymbol{a}_t)$.

Let us now define by 
\begin{align*}
\frak{L}_t(s_{\tau_k},\boldsymbol{a}):=\max\left\{\mathcal{M}^{\boldsymbol{\pi}}Q_{2,t}(s_{\tau_k},\boldsymbol{a}), R^\explore(s_{\tau_k},\boldsymbol{a})+\gamma\underset{a'\in\mathcal{A}}{\max}\;Q_{2,t}(s',\boldsymbol{a'})\right\}-Q_2^\star(s_t,a).
\end{align*}
Then
\begin{align}
\Xi_{2,t+1}(s_t,\boldsymbol{a}_t)=(1-\alpha_t(s_t,\boldsymbol{a}_t))\Xi_{2,t}(s_t,\boldsymbol{a}_t)+\alpha_t(s_t,\boldsymbol{a}_t))\left[\frak{L}_t(s_{\tau_k},a)\right].   
\end{align}

We now observe that
\begin{align}\nonumber
\mathbb{E}\left[\frak{L}_t(s_{\tau_k},\boldsymbol{a})|\mathcal{F}_t\right]&=\sum_{s'\in\mathcal{S}}P(s';\boldsymbol{a},s_{\tau_k})\max\left\{\mathcal{M}^{\boldsymbol{\pi}}Q_2(s_{\tau_k},\boldsymbol{a}), R^\explore(s_{\tau_k},\boldsymbol{a})+\gamma\underset{a'\in\mathcal{A}}{\max}\;Q_{2,t}(s',\boldsymbol{a'})\right\}-Q_2^\star(s_{\tau_k},\boldsymbol{a})
\\&= T Q_{2,t}(s,\boldsymbol{a})-Q_2^\star(s,\boldsymbol{a}). \label{expectation_L}
\end{align}
Now, using the fixed point property that implies $Q_2^\star=T Q_2^\star$, we find that
\begin{align}\nonumber
    \mathbb{E}\left[\frak{L}_t(s_{\tau_k},\boldsymbol{a})|\mathcal{F}_t\right]&=T Q_{2,t}(s,\boldsymbol{a})-T Q_2^\star(s,\boldsymbol{a})
    \\&\leq\left\|T Q_{2,t}-T Q_2^\star\right\|\nonumber
    \\&\leq \gamma\left\| Q_{2,t}- Q_2^\star\right\|_\infty=\gamma\left\|\Xi_t\right\|_\infty.
\end{align}
using the contraction property of $T$ established in Lemma \ref{lemma:bellman_contraction}. This proves (ii).

We now prove iii), that is
\begin{align}
    {\rm Var}\left[L_t|\mathcal{F}_t\right]\leq c(1+\|\Xi_t\|^2).
\end{align}
Now by \eqref{expectation_L} we have that
\begin{align*}
  {\rm Var}\left[L_t|\mathcal{F}_t\right]&= {\rm Var}\left[\max\left\{\mathcal{M}^{\boldsymbol{\pi}}Q_2(s_{\tau_k},\boldsymbol{a}), R^\explore(s_{\tau_k},\boldsymbol{a})+\gamma\underset{a'\in\mathcal{A}}{\max}\;Q_{2,t}(s',\boldsymbol{a'})\right\}-Q_2^\star(s_t,\boldsymbol{a})\right]
  \\&= \mathbb{E}\Bigg[\Bigg(\max\left\{\mathcal{M}^{\boldsymbol{\pi}}Q_2(s_{\tau_k},\boldsymbol{a}), R^\explore(s_{\tau_k},\boldsymbol{a})+\gamma\underset{a'\in\mathcal{A}}{\max}\;Q_{2,t}(s',\boldsymbol{a'})\right\}
  \\&\qquad\qquad\qquad\qquad\qquad\quad\quad\quad-Q_2^\star(s_t,a)-\left(T Q_{2,t}(s,\boldsymbol{a})-Q_2^\star(s,\boldsymbol{a})\right)\Bigg)^2\Bigg]
      \\&= \mathbb{E}\left[\left(\max\left\{\mathcal{M}^{\boldsymbol{\pi}}Q_2(s_{\tau_k},\boldsymbol{a}), R^\explore(s_{\tau_k},\boldsymbol{a})+\gamma\underset{a'\in\mathcal{A}}{\max}\;Q_{2,t}(s',\boldsymbol{a'})\right\}-T Q_{2,t}(s,\boldsymbol{a})\right)^2\right]
    \\&= {\rm Var}\left[\max\left\{\mathcal{M}^{\boldsymbol{\pi}}Q_2(s_{\tau_k},\boldsymbol{a}), R^\explore(s_{\tau_k},\boldsymbol{a})+\gamma\underset{a'\in\mathcal{A}}{\max}\;Q_{2,t}(s',\boldsymbol{a'})\right\}-T Q_{2,t}(s,\boldsymbol{a}))^2\right]
    \\&\leq c(1+\|\Xi_t\|^2),
\end{align*}
for some $c>0$ where the last line follows due to the boundedness of $Q_2$ (which follows from Assumptions 2 and 4). 

\end{proof}
This concludes the proof of Prop. \ref{theorem:q_learning}.

\section*{Proof of Proposition \ref{prop:switching_times}}
\begin{proof}[Proof of Prop. \ref{prop:switching_times}]
The proof is given by establishing a contradiction. Therefore suppose that $\mathcal{M}^{\pi,\pi^\explore}v^{\pi^\exploit,\pi'^\explore}_2(s_{\tau_k})\leq v^{\pi^\exploit,\pi'^\explore}_2(s_{\tau_k})$ and suppose that the intervention time $\tau'_1>\tau_1$ is an optimal intervention time. Construct the policy $\pi'^\explore\in\Pi^\explore$ and $\tilde{\pi}^\explore$ policy switching times by $(\tau'_0,\tau'_1,\ldots,)$ and $\pi'^\explore\in\Pi^\explore$ policy by $(\tau'_0,\tau_1,\ldots)$ respectively.  Define by $l=\inf\{t>0;\mathcal{M}^{\pi,\pi^\explore}v^{\pi^\exploit,\pi'^\explore}_2(s_{t})= v^{\pi^\exploit,\pi'^\explore}_2(s_{t})\}$ and $m=\sup\{t;t<\tau'_1\}$.
By construction we have that
\begin{align*}
& \quad v^{\pi^\exploit,\pi'^\explore}_2(s)
\\&=\mathbb{E}\left[-L(s_{0},a_{0})+\mathbb{E}\left[\ldots+\gamma^{l-1}\mathbb{E}\left[-L(s_{\tau_1-1},a_{\tau_1-1})+\ldots+\gamma^{m-l-1}\mathbb{E}\left[ -L(s_{\tau'_1-1},a_{\tau'_1-1})+\gamma\mathcal{M}^{\pi^\exploit,\pi'^\explore}v^{\pi^\exploit,\pi'^\explore}_2(s')\right]\right]\right]\right]
\\&<\mathbb{E}\left[-L(s_{0},a_{0})+\mathbb{E}\left[\ldots+\gamma^{l-1}\mathbb{E}\left[ -L(s_{\tau_1-1},a_{\tau_1-1})+\gamma\mathcal{M}^{\pi^\exploit,\tilde{\pi}^\explore}v^{\pi^\exploit,\pi'^\explore}_2(s_{\tau_1})\right]\right]\right]
\end{align*}
We now use the following observation $\mathbb{E}\left[ -L(s_{\tau_1-1},a_{\tau_1-1})+\gamma\mathcal{M}^{\pi^\exploit,\tilde{\pi}^\explore}v^{\pi^\exploit,\pi'^\explore}_2(s_{\tau_1})\right]\\\ \text{\hspace{30 mm}}\leq \max\left\{\mathcal{M}^{\pi^\exploit,\tilde{\pi}^\explore}v^{\pi^\exploit,\pi'^\explore}_2(s_{\tau_1}),\underset{a_{\tau_1}\in\mathcal{A}}{\max}\;\left[ -L(s_{\tau_{1}},a_{\tau_{1}})+\gamma\sum_{s'\in\mathcal{S}}P(s';a_{\tau_1},s_{\tau_1})v^{\pi^\exploit,\pi^\explore}_2(s')\right]\right\}$.

Using this we deduce that
\begin{align*}
&v^{\pi^\exploit,\pi'^\explore}_2(s)\leq\mathbb{E}\Bigg[-L(s_{0},a_{0})+\mathbb{E}\Bigg[\ldots
\\&+\gamma^{l-1}\mathbb{E}\left[ -L(s_{\tau_1-1},a_{\tau_1-1})+\gamma\max\left\{\mathcal{M}^{\pi^\exploit,\tilde{\pi}^\explore}v^{\pi^\exploit,\pi'^\explore}_2(s_{\tau_1}),\underset{a_{\tau_1}\in\mathcal{A}}{\max}\;\left[ -L(s_{\tau_{1}},a_{\tau_{1}})+\gamma\sum_{s'\in\mathcal{S}}P(s';a_{\tau_1},s_{\tau_1})v^{\pi^\exploit,\pi^\explore}_2(s')\right]\right\}\right]\Bigg]\Bigg]
\\&=\mathbb{E}\left[-L(s_{0},a_{0})+\mathbb{E}\left[\ldots+\gamma^{l-1}\mathbb{E}\left[ -L(s_{\tau_1-1},a_{\tau_1-1})+\gamma\left[T v^{\pi^\exploit,\tilde{\pi}^\explore}_2\right](s_{\tau_1})\right]\right]\right]=v^{\pi^\exploit,\tilde{\pi}^\explore}_2(s)),
\end{align*}
where the first inequality is true by assumption on $\mathcal{M}$. This is a contradiction since $\pi'^\explore$ is an optimal policy for Player 2. Using analogous reasoning, we deduce the same result for $\tau'_k<\tau_k$ after which deduce the result. Moreover, by invoking the same reasoning, we can conclude that it must be the case that $(\tau_0,\tau_1,\ldots,\tau_{k-1},\tau_k,\tau_{k+1},\ldots,)$ are the optimal switching times, this completes the proof of part (i).

We now prove part (ii). First, we note that it is easy to see that $v^{\pi^\exploit,(\pi^\explore,\mathfrak{g})}_2$ is bounded above, indeed using the above we have that

\begin{align}
v^{\pi^\exploit,(\pi^\explore,\mathfrak{g})}_2(s)&=\mathbb{E}\left[\sum_{t\geq 0}\gamma^t \left(R^\explore\left(s_t,a^\exploit_t,a^\explore_t\right)  - \boldsymbol{1}_{\mathcal{A}^\explore}(a^\explore_t)\right)\right]
\\&=\mathbb{E}\left[ \sum_{t=0}^\infty \gamma^t\left(-\Big(L(s,a^\explore_t)\boldsymbol{1}_{\mathcal{A}^\explore}(a^\explore_t)+L(s_t,a^\exploit)(1-\boldsymbol{1}_{\mathcal{A}^\explore}(a^\explore_t))\Big)- \boldsymbol{1}_{\mathcal{A}^\explore}(a^\explore_t)\right)\right]
\\&\leq \left|\mathbb{E}_{\pi,\pi^\explore}\left[ \sum_{t=0}^\infty \gamma^t\left(2{L}- \boldsymbol{1}_{\mathcal{A}^\explore}(a^\explore_t)\right)\right]\right|
\\&\leq  \sum_{t=0}^\infty \gamma^t\left(2\left\|{L}\right\| +K\right)
\\&=\frac{1}{1-\gamma}\left(2\left\|{L}\right\| +K\right),
\end{align}
using the triangle inequality,  the (upper-)boundedness of $L$ (Assumption 5).
We now note that by the dominated convergence theorem we have that $\forall (s_0)\in\mathcal{S}\times\{0,1\}$
\begin{align}
&\underset{l\to0}{\lim}\; v^{\pi^\exploit,(\pi^\explore,\mathfrak{g})}_2(s)  = \underset{l\to 0}{\lim}\; \mathbb{E}\left[\sum_{t\geq 0}\gamma^t \left(R^\explore\left(s_t,a^\exploit_t,a^\explore_t\right)  - \boldsymbol{1}_{\mathcal{A}^\explore}(a^\explore_t)\right)\right]
\\&=\underset{l\to 0}{\lim}\;\mathbb{E}\left[ \sum_{t=0}^\infty \gamma^t\left(-L(s,a^\explore_t)\boldsymbol{1}_{\mathcal{A}^\explore}(a^\explore_t)-L(s_t,a^\exploit)(1-\boldsymbol{1}_{\mathcal{A}^\explore}(a^\explore_t))- \boldsymbol{1}_{\mathcal{A}^\explore}(a^\explore_t)\right)\right]
\\&=\mathbb{E}\underset{l\to 0}{\lim}\;\left[ \sum_{t=0}^\infty \gamma^t\left(-L(s,a^\explore_t)\boldsymbol{1}_{\mathcal{A}^\explore}(a^\explore_t)-L(s_t,a^\exploit)(1-\boldsymbol{1}_{\mathcal{A}^\explore}(a^\explore_t))- \boldsymbol{1}_{\mathcal{A}^\explore}(a^\explore_t)\right)\right]
\\&=-\mathbb{E}\left[ \sum_{t=0}^\infty \gamma^t \boldsymbol{1}_{\mathcal{A}^\explore}(a^\explore_t)\right],\label{prop_1_2_last_exp}
\end{align}
using Assumption 6 in the last step, after which we deduce (ii) since \eqref{prop_1_2_last_exp} is maximised when $\boldsymbol{1}_{\mathcal{A}^\explore}(a^\explore_t)=0$ for all $t=0,1,\ldots$ which is achieved only when $\mu_{l}(\mathfrak{g})=0$. Additionally, by part (i) we have that 
\begin{align}
\tau_k=\inf\left\{\tau>\tau_{k-1}|\mathcal{M}^{\Pi^\explore}v_2^{\pi^\exploit,\Pi^\explore}= v_2^{\pi^\exploit,\Pi^\explore}\right\}.\label{obstacle_problem_app}
\end{align}
It is easy to see that given \eqref{prop_1_2_last_exp} and the definition of $\mathcal{M}$ (c.f. \eqref{intervention_op}), condition \eqref{obstacle_problem_app} can never be satisfied which implies that \switcher\; performs no interventions. 

This completes the proof of Prop. \ref{prop:switching_times}.
\end{proof}

To complete the proof of Theorem \ref{convergence_theorem}, we prove the following result:

\begin{lemma}\label{lemma:explorer_convergence}
The {\fontfamily{cmss}\selectfont Explorer} learns to solve the MDP $\left\langle \mathcal{S},\mathcal{A},P,R,\gamma\right\rangle$ and its value function converges.
\end{lemma}
\begin{proof}
We first deduce the boundedness of the {\fontfamily{cmss}\selectfont Exploiter} objective $v^{\pi^\exploit,(\pi^\explore,\mathfrak{g})}_1$:
\begin{align*}
v^{\pi^\exploit,(\pi^\explore,\mathfrak{g})}_1(s)&=\mathbb{E}\left[\sum_{t\geq 0}\gamma_{1}^tR^\exploit\left(s_t,a^\exploit_t,a^\explore_t\right)\right]    
\\&=\mathbb{E}\left[\sum_{t\geq 0}\gamma_{1}^t\left(R(s_t,a_t^\exploit)(1-\beta\boldsymbol{1}_{\mathcal{A}^\explore}(a^\explore_t)) + R(s,a^\explore_t)\beta\boldsymbol{1}_{\mathcal{A}^\explore}(a^\explore_t)\right)\right]
\\&\leq\mathbb{E}\left[\sum_{t\geq 0}\gamma_{1}^t\left(R(s_t,a_t^\exploit) + R(s,a^\explore_t)\right)\right]
\\&\leq \frac{2}{1-\gamma_1}\|R\|,
\end{align*}
therefore the {\fontfamily{cmss}\selectfont Explorer}'s objective is bounded above by some finite quantity.

Using the kronecker-delta function, the {\fontfamily{cmss}\selectfont Exploiter} objective as:
\begin{align}
&v^{\pi^\exploit,(\pi^\explore,\mathfrak{g})}_1(s)=\mathbb{E}\left[\sum_{t\geq 0}\sum_{k= 0}^{\mu_{l}(\mathfrak{g})}\gamma_{1}^tR(s_t,a_t^\exploit)(1-\delta^t_{\tau_k}) + R(s,a^\explore_t)\delta^t_{\tau_k}\right]. 
\end{align}

Recall that $ \mu_{l}(\mathfrak{g}) $ denotes the number of switch activations performed by the {\fontfamily{cmss}\selectfont Switcher}. Denote by $\mathfrak{g}_0$ the {\fontfamily{cmss}\selectfont Explorer} intervention policy that performs no interventions. By Prop. \ref{prop:switching_times} and by the dominated convergence theorem, we have that $\forall s\in\cS$
\begin{align*}
    \underset{l\to0}{\lim}v^{\pi^\exploit,(\pi^\explore,\mathfrak{g})}_1(s)&=\underset{l\to0}{\lim}\mathbb{E}_{P,\pi^\exploit,(\pi^\explore,\mathfrak{g})}\left[\sum_{t\geq 0}\sum_{k= 0}^{\mu_{l}(\mathfrak{g})}\gamma_{1}^tR(s_t,a_t^\exploit)(1-\delta^t_{\tau_k}) + R(s,a^\explore_t)\delta^t_{\tau_k}\right]
    \\&=\mathbb{E}_{P,\pi^\exploit,(\pi^\explore,\mathfrak{g})}\underset{l\to0}{\lim}\left[\sum_{t\geq 0}\sum_{k= 0}^{\mu_{l}(\mathfrak{g})}\gamma_{1}^tR(s_t,a_t^\exploit)(1-\delta^t_{\tau_k}) + R(s,a^\explore_t)\delta^t_{\tau_k}\right]
        \\&=\mathbb{E}_{P,\pi^\exploit,(\pi^\explore,\mathfrak{g}_0)}\left[\sum_{t\geq 0}\gamma_{1}^tR(s_t,a_t^\exploit)\right]
        \\&=\mathbb{E}_{P,\pi^\exploit}\left[\sum_{t\geq 0}\gamma_{1}^tR(s_t,a_t^\exploit)\right]=v^{\pi^\exploit}_1(s),
\end{align*}
using the fact that $\underset{l\to0}{\lim}\;\mu_{l}=0$ and by Fubini's theorem in the penultimate step.  Therefore, in the limit $l\to0$, the {\fontfamily{cmss}\selectfont Exploiter} solves the MDP $\left\langle \mathcal{S},\mathcal{A},P,R,\gamma\right\rangle$ which converges to a stable point.
\begin{corollary}
After combining Lemma \ref{lemma:explorer_convergence}, Prop.  \ref{prop:switching_times} and Prop. \ref{theorem:q_learning} we deduce the result of Theorem \ref{convergence_theorem}.
\end{corollary}
\end{proof}
\section*{Proof of Convergence with Function Approximation}
First let us recall the statement of the theorem:
\begin{customthm}{3}
SEREN converges to a limit point $r^\star$ which is the unique solution to the equation:
\begin{align}
\Pi \mathfrak{F} (\Phi r^\star)=\Phi r^\star, \qquad \text{a.e.}
\end{align}
where we recall that for any test function $\Lambda \in \mathcal{V}$, the operator $\mathfrak{F}$ is defined by $
    \mathfrak{F}\Lambda:=\Theta+\gamma P \max\{\mathcal{M}\Lambda,\Lambda\}$.

Moreover, $r^\star$ satisfies the following:
\begin{align}
    \left\|\Phi r^\star - Q_2^\star\right\|\leq c\left\|\Pi Q_2^\star-Q_2^\star\right\|.
\end{align}
\end{customthm}

The theorem is proven using a set of results that we now establish. To this end, we first wish to prove the following bound:    
\begin{lemma}
For any $Q\in\mathcal{V}$ we have that
\begin{align}
    \left\|\mathfrak{F}Q_2-Q'_2\right\|\leq \gamma\left\|Q_2-Q'_2\right\|,
\end{align}
so that the operator $\mathfrak{F}$ is a contraction.
\end{lemma}
\begin{proof}
Recall, for any test function $\psi$ , a projection operator $\Pi$ acting $\Lambda$ is defined by the following 
\begin{align*}
\Pi \Lambda:=\underset{\bar{\Lambda}\in\{\Phi r|r\in\mathbb{R}^p\}}{\arg\min}\left\|\bar{\Lambda}-\Lambda\right\|. 
\end{align*}
Now, we first note that in the proof of Lemma \ref{lemma:bellman_contraction}, we deduced that for any $\Lambda\in L_2$ we have that
\begin{align*}
    \left\|\mathcal{M}^{\boldsymbol{\pi}}\Lambda-\left[ \psi(\cdot,\boldsymbol{a})+\gamma\underset{\boldsymbol{a}\in\boldsymbol{\mathcal{A}}}{\max}\;\mathcal{P}^{\boldsymbol{a}}\Lambda'\right]\right\|\leq \gamma\left\|\Lambda-\Lambda'\right\|,
\end{align*}
(c.f. Lemma \ref{lemma:bellman_contraction}). 

Setting $\Lambda=Q_2$ and $\psi=\cR$, it can be straightforwardly deduced that for any $Q_2,\hat{Q}_2\in L_2$:
    $\left\|\mathcal{M}^{\boldsymbol{\pi}}Q_2-\hat{Q}_2\right\|\leq \gamma\left\|Q_2-\hat{Q}_2\right\|$. Hence, using the contraction property of $\mathcal{M}$, we readily deduce the following bound:
\begin{align}\max\left\{\left\|\mathcal{M}^{\boldsymbol{\pi}}Q_2-\hat{Q}_2\right\|,\left\|\mathcal{M}^{\boldsymbol{\pi}}Q_2-\mathcal{M}\hat{Q}_2\right\|\right\}\leq \gamma\left\|Q_2-\hat{Q}_2\right\|,
\label{m_bound_q_twice}
\end{align}
    
We now observe that $\mathfrak{F}$ is a contraction. Indeed, since for any $Q_2,Q'_2\in L_2$ we have that:
%
%
%
\begin{align*}
\left\|\mathfrak{F}Q_2-\mathfrak{F}Q_2'\right\|&=\left\|\Theta+\gamma P \max\{\mathcal{M}^{\boldsymbol{\pi}}Q_2,Q_2\}-\left(\Theta+\gamma P \max\{\mathcal{M}^{\boldsymbol{\pi}}Q'_2,Q'_2\}\right)\right\|
\\&=\gamma \left\|P \max\{\mathcal{M}^{\boldsymbol{\pi}}Q_2,Q_2\}-P \max\{\mathcal{M}^{\boldsymbol{\pi}}Q'_2,Q'_2\}\right\|
\\&\leq\gamma \left\| \max\{\mathcal{M}^{\boldsymbol{\pi}}Q_2,Q_2\}- \max\{\mathcal{M}^{\boldsymbol{\pi}}Q'_2,Q'_2\}\right\|
\\&\leq\gamma \left\| \max\{\mathcal{M}^{\boldsymbol{\pi}}Q_2-\mathcal{M}^{\boldsymbol{\pi}}Q'_2,Q_2-\mathcal{M}^{\boldsymbol{\pi}}Q'_2,\mathcal{M}^{\boldsymbol{\pi}}Q_2-Q'_2,Q_2-Q'_2\}\right\|
\\&\leq\gamma \max\{\left\|\mathcal{M}^{\boldsymbol{\pi}}Q_2-\mathcal{M}^{\boldsymbol{\pi}}Q'_2\right\|,\left\|Q_2-\mathcal{M}^{\boldsymbol{\pi}}Q'_2\right\|,\left\|\mathcal{M}^{\boldsymbol{\pi}}Q_2-Q'_2\right\|,\left\|Q_2-Q'_2\right\|\}
\\&=\gamma\left\|Q_2-Q'_2\right\|,
\end{align*}
using \eqref{m_bound_q_twice} and again using the non-expansiveness of $P$.
\end{proof}
We next show that the following two bounds hold:
\begin{lemma}\label{projection_F_contraction_lemma}
For any $Q_2\in\mathcal{V}$ we have that
\begin{itemize}
    \item[i)] 
$\qquad\qquad
    \left\|\Pi \mathfrak{F}Q_2-\Pi \mathfrak{F}\bar{Q}_2\right\|\leq \gamma\left\|Q_2-\bar{Q}_2\right\|$,
    \item[ii)]$\qquad\qquad\left\|\Phi r^\star - Q_2^\star\right\|\leq \frac{1}{\sqrt{1-\gamma^2}}\left\|\Pi Q_2^\star - Q_2^\star\right\|$. 
\end{itemize}
\end{lemma}
\begin{proof}
The first result is straightforward since as $\Pi$ is a projection it is non-expansive and hence:
\begin{align*}
    \left\|\Pi \mathfrak{F}Q_2-\Pi \mathfrak{F}\bar{Q}_2\right\|\leq \left\| \mathfrak{F}Q_2-\mathfrak{F}\bar{Q}_2\right\|\leq \gamma \left\|Q_2-\bar{Q}_2\right\|,
\end{align*}
using the contraction property of $\mathfrak{F}$. This proves i). For ii), we note that by the orthogonality property of projections we have that $\left\langle\Phi r^\star - \Pi Q_2^\star,\Phi r^\star - \Pi Q_2^\star\right\rangle$, hence we observe that:
\begin{align*}
    \left\|\Phi r^\star - Q_2^\star\right\|^2&=\left\|\Phi r^\star - \Pi Q_2^\star\right\|^2+\left\|\Phi r^\star - \Pi Q_2^\star\right\|^2
\\&=\left\|\Pi \mathfrak{F}\Phi r^\star - \Pi Q_2^\star\right\|^2+\left\|\Phi r^\star - \Pi Q_2^\star\right\|^2
\\&\leq\left\|\mathfrak{F}\Phi r^\star -  Q_2^\star\right\|^2+\left\|\Phi r^\star - \Pi Q_2^\star\right\|^2
\\&=\left\|\mathfrak{F}\Phi r^\star -  \mathfrak{F}Q_2^\star\right\|^2+\left\|\Phi r^\star - \Pi Q_2^\star\right\|^2
\\&\leq\gamma^2\left\|\Phi r^\star -  Q_2^\star\right\|^2+\left\|\Phi r^\star - \Pi Q_2^\star\right\|^2,
\end{align*}
after which we readily deduce the desired result.
\end{proof}

\begin{lemma}
Define  the operator $H$ by the following: $
  HQ_2(z)=  \begin{cases}
			\mathcal{M}^{\boldsymbol{\pi}}Q_2(z), & \text{if $\mathcal{M}^{\boldsymbol{\pi}}Q_2(s)>\Phi r^\star,$}\\
            Q_2(z), & \text{otherwise},
		 \end{cases}$
\\and $\tilde{\mathfrak{F}}$ by: $
    \tilde{\mathfrak{F}}Q_2:=\cR +\gamma PHQ_2$.

For any $Q_2,\bar{Q}_2\in L_2$ we have that
\begin{align}
    \left\|\tilde{\mathfrak{F}}Q_2-\tilde{\mathfrak{F}}\bar{Q}_2\right\|\leq \gamma \left\|Q_2-\bar{Q}_2\right\|
\end{align}
and hence $\tilde{\mathfrak{F}}$ is a contraction mapping.
\end{lemma}
\begin{proof}
Using \eqref{m_bound_q_twice}, we now observe that
\begin{align*}
    \left\|\tilde{\mathfrak{F}}Q_2-\tilde{\mathfrak{F}}\bar{Q}_2\right\|&=\left\|\cR+\gamma PHQ_2 -\left(\cR+\gamma PH\bar{Q}_2\right)\right\|
\\&\leq \gamma\left\|HQ_2 - H\bar{Q}_2\right\|
\\&\leq \gamma\left\|\max\left\{\mathcal{M}^{\boldsymbol{\pi}}Q_2-\mathcal{M}\bar{Q}_2,Q_2-\bar{Q}_2,\mathcal{M}^{\boldsymbol{\pi}}Q_2-\bar{Q}_2,\mathcal{M}\bar{Q}_2-Q_2\right\}\right\|
\\&\leq \gamma\max\left\{\left\|\mathcal{M}^{\boldsymbol{\pi}}Q_2-\mathcal{M}\bar{Q}_2\right\|,\left\|Q_2-\bar{Q}_2\right\|,\left\|\mathcal{M}^{\boldsymbol{\pi}}Q_2-\bar{Q}_2\right\|,\left\|\mathcal{M}\bar{Q}_2-Q_2\right\|\right\}
\\&\leq \gamma\max\left\{\gamma\left\|Q_2-\bar{Q}_2\right\|,\left\|Q_2-\bar{Q}_2\right\|,\left\|\mathcal{M}^{\boldsymbol{\pi}}Q_2-\bar{Q}_2\right\|,\left\|\mathcal{M}\bar{Q}_2-Q_2\right\|\right\}
\\&=\gamma\left\|Q_2-\bar{Q}_2\right\|,
\end{align*}
again using the non-expansive property of $P$.
\end{proof}
\begin{lemma}
Define by $\tilde{Q}_2:=R^\explore+\gamma Pv^{\boldsymbol{\tilde{\pi}}}_2$ where
\begin{align}
    v^{\boldsymbol{\tilde{\pi}}}_2(s):= R_2(s_{\tau_k},a)+\gamma\underset{a\in\mathcal{A}}{\max}\;\sum_{s'\in\mathcal{S}}P(s';a,s_{\tau_k})\Phi r^\star(s'), \label{v_tilde_definition}
\end{align}
then $\tilde{Q}_2$ is a fixed point of $\tilde{\mathfrak{F}}\tilde{Q}_2$, that is $\tilde{\mathfrak{F}}\tilde{Q}_2=\tilde{Q}_2$. 
\end{lemma}
\begin{proof}
We begin by observing that
\begin{align*}
H\tilde{Q}_2(z)&=H\left(L(z)+\gamma Pv_2^{\boldsymbol{\tilde{\pi}}}\right)    
\\&= \begin{cases}
			\mathcal{M}^{\boldsymbol{\pi}}Q_2(z), & \text{if $\mathcal{M}^{\boldsymbol{\pi}}Q_2(z)>\Phi r^\star,$}\\
            Q_2(z), & \text{otherwise},
		 \end{cases}
\\&= \begin{cases}
			\mathcal{M}^{\boldsymbol{\pi}}Q_2(z), & \text{if $\mathcal{M}^{\boldsymbol{\pi}}Q_2(z)>\Phi r^\star,$}\\
            L(z)+\gamma Pv_2^{\boldsymbol{\tilde{\pi}}}, & \text{otherwise},
		 \end{cases}
\\&=v_2^{\boldsymbol{\tilde{\pi}}}(s).
\end{align*}
Hence,
\begin{align}
    \tilde{\mathfrak{F}}\tilde{Q}_2=\cR+\gamma PH\tilde{Q}_2=\cR+\gamma Pv_2^{\boldsymbol{\tilde{\pi}}}=\tilde{Q}_2. 
\end{align}
which proves the result.
\end{proof}
\begin{lemma}\label{value_difference_Q_difference}
The following bound holds:
\begin{align}
    \mathbb{E}\left[v_2^{\boldsymbol{\hat{\pi}}}(s_0)\right]-\mathbb{E}\left[v_2^{\boldsymbol{\tilde{\pi}}}(s_0)\right]\leq 2\left[(1-\gamma)\sqrt{(1-\gamma^2)}\right]^{-1}\left\|\Pi Q_2^\star -Q_2^\star\right\|.
\label{F_tilde_fixed_point}\end{align}
\end{lemma}
\begin{proof}

By definitions of $v_2^{\boldsymbol{\hat{\pi}}}$ and $v_2^{\boldsymbol{\tilde{\pi}}}$ (c.f \eqref{v_tilde_definition}) and using Jensen's inequality and the stationarity property we have that,
\begin{align}\nonumber
    \mathbb{E}\left[v_2^{\boldsymbol{\hat{\pi}}}(s_0)\right]-\mathbb{E}\left[v_2^{\boldsymbol{\tilde{\pi}}}(s_0)\right]&=\mathbb{E}\left[Pv_2^{\boldsymbol{\hat{\pi}}}(z_0)\right]-\mathbb{E}\left[Pv_2^{\boldsymbol{\tilde{\pi}}}(z_0)\right]
    \\&\leq \left|\mathbb{E}\left[Pv_2^{\boldsymbol{\hat{\pi}}}(z_0)\right]-\mathbb{E}\left[Pv_2^{\boldsymbol{\tilde{\pi}}}(z_0)\right]\right|\nonumber
    \\&\leq \left\|Pv_2^{\boldsymbol{\hat{\pi}}}-Pv_2^{\boldsymbol{\tilde{\pi}}}\right\|. \label{v_approx_intermediate_bound_P}
\end{align}
Now recall that $\tilde{Q}_2:=\cR+\gamma Pv_2^{\boldsymbol{\tilde{\pi}}}$ and $Q_2^\star:=\cR+\gamma Pv_2^{\boldsymbol{\pi^\star}}$,  using these expressions in \eqref{v_approx_intermediate_bound_P} we find that 
\begin{align*}
    \mathbb{E}\left[v_2^{\boldsymbol{\hat{\pi}}}(z_0)\right]-\mathbb{E}\left[v_2^{\boldsymbol{\tilde{\pi}}}(z_0)\right]&\leq \frac{1}{\gamma}\left\|\tilde{Q}_2-Q_2^\star\right\|. \label{v_approx_q_approx_bound}
\end{align*}
Moreover, by the triangle inequality and using the fact that $\mathfrak{F}(\Phi r^\star)=\tilde{\mathfrak{F}}(\Phi r^\star)$ and that $\mathfrak{F}Q_2^\star=Q_2^\star$ and $\mathfrak{F}\tilde{Q}_2=\tilde{Q}_2$ (c.f. \eqref{F_tilde_fixed_point}) we have that
\begin{align*}
\left\|\tilde{Q}_2-Q_2^\star\right\|&\leq \left\|\tilde{Q}_2-\mathfrak{F}(\Phi r^\star)\right\|+\left\|Q_2^\star-\tilde{\mathfrak{F}}(\Phi r^\star)\right\|    
\\&\leq \gamma\left\|\tilde{Q}_2-\Phi r^\star\right\|+\gamma\left\|Q_2^\star-\Phi r^\star\right\| 
\\&\leq 2\gamma\left\|\tilde{Q}_2-\Phi r^\star\right\|+\gamma\left\|Q_2^\star-\tilde{Q}_2\right\|, 
\end{align*}
which gives the following bound:
\begin{align*}
\left\|\tilde{Q}_2-Q_2^\star\right\|&\leq 2\left(1-\gamma\right)^{-1}\left\|\tilde{Q}_2-\Phi r^\star\right\|, 
\end{align*}
from which, using Lemma \ref{projection_F_contraction_lemma}, we deduce that $
    \left\|\tilde{Q}_2-Q_2^\star\right\|\leq 2\left[(1-\gamma)\sqrt{(1-\gamma^2)}\right]^{-1}\left\|\tilde{Q}_2-\Phi r^\star\right\|$,
after which by \eqref{v_approx_q_approx_bound}, we finally obtain
\begin{align*}
        \mathbb{E}\left[v_2^{\boldsymbol{\hat{\pi}}}(s_0)\right]-\mathbb{E}\left[v_2^{\boldsymbol{\tilde{\pi}}}(s_0)\right]\leq  2\left[(1-\gamma)\sqrt{(1-\gamma^2)}\right]^{-1}\left\|\tilde{Q}_2-\Phi r^\star\right\|,
\end{align*}
as required.
\end{proof}

Let us rewrite the update in the following way:
\begin{align*}
    r_{t+1}=r_t+\gamma_t\Xi_2(w_t,r_t),
\end{align*}
where the function $\Xi_2:\mathbb{R}^{2d}\times \mathbb{R}^p\to\mathbb{R}^p$ is given by:
\begin{align*}
\Xi_2(w,r):=\phi(z)\left(L(z)+\gamma\max\left\{(\Phi r) (z'),\mathcal{M}(\Phi r) (z')\right\}-(\Phi r)(s)\right),
\end{align*}
for any $w\equiv (z,z')\in\left(\mathbb{N}\times\mathcal{S}\right)^2$ where $z=(t,s)\in\mathbb{N}\times\mathcal{S}$ and $z'=(t,s')\in\mathbb{N}\times\mathcal{S}$  and for any $r\in\mathbb{R}^p$. Let us also define the function $\boldsymbol{\Xi}_2:\mathbb{R}^p\to\mathbb{R}^p$ by the following:
\begin{align*}
    \boldsymbol{\Xi}_2(r):=\mathbb{E}_{w_0\sim (\mathbb{P},\mathbb{P})}\left[\Xi_2(w_0,r)\right]; w_0:=(z_0,z_1).
\end{align*}
\begin{lemma}\label{iteratation_property_lemma}
The following statements hold for all $z\in \{0,1\}\times \mathcal{S}$:
\begin{itemize}
    \item[i)] $
(r-r^\star)\boldsymbol{\Xi}_{2,k}(r)<0,\qquad \forall r\neq r^\star,    
$
\item[ii)] $
\boldsymbol{\Xi}_{2,k}(r^\star)=0$.
\end{itemize}
\end{lemma}
\begin{proof}
To prove the statement, we first note that each component of $\boldsymbol{\Xi}_{2,k}(r)$ admits a representation as an inner product, indeed: 
\begin{align*}
\boldsymbol{\Xi}_{2,k}(r)&=\mathbb{E}\left[\phi_k(z_0)(L(z_0)+\gamma\max\left\{\Phi r(z_1),\mathcal{M}^{\boldsymbol{\pi}}\Phi(z_1)\right\}-(\Phi r)(z_0)\right] 
\\&=\mathbb{E}\left[\phi_k(z_0)(L(z_0)+\gamma\mathbb{E}\left[\max\left\{\Phi r(z_1),\mathcal{M}^{\boldsymbol{\pi}}\Phi(z_1)\right\}|z_0\right]-(\Phi r)(z_0)\right]
\\&=\mathbb{E}\left[\phi_k(z_0)(L(z_0)+\gamma P\max\left\{\left(\Phi r,\mathcal{M}^{\boldsymbol{\pi}}\Phi\right)\right\}(z_0)-(\Phi r)(z_0)\right]
\\&=\left\langle\phi_k,\mathfrak{F}\Phi r-\Phi r\right\rangle,
\end{align*}
using the iterated law of expectations and the definitions of $P$ and $\mathfrak{F}$.

We now are in position to prove i). Indeed, we now observe the following:
\begin{align*}
\left(r-r^\star\right)\boldsymbol{\Xi}_{2,k}(r)&=\sum_{l=1}\left(r(l)-r^\star(l)\right)\left\langle\phi_l,\mathfrak{F}\Phi r -\Phi r\right\rangle
\\&=\left\langle\Phi r -\Phi r^\star, \mathfrak{F}\Phi r -\Phi r\right\rangle
\\&=\left\langle\Phi r -\Phi r^\star, (\boldsymbol{1}-\Pi)\mathfrak{F}\Phi r+\Pi \mathfrak{F}\Phi r -\Phi r\right\rangle
\\&=\left\langle\Phi r -\Phi r^\star, \Pi \mathfrak{F}\Phi r -\Phi r\right\rangle,
\end{align*}
where in the last step we used the orthogonality of $(\boldsymbol{1}-\Pi)$. We now recall that $\Pi \mathfrak{F}\Phi r^\star=\Phi r^\star$ since $\Phi r^\star$ is a fixed point of $\Pi \mathfrak{F}$. Additionally, using Lemma \ref{projection_F_contraction_lemma} we observe that $\|\Pi \mathfrak{F}\Phi r -\Phi r^\star\| \leq \gamma \|\Phi r -\Phi r^\star\|$. With this we now find that
\begin{align*}
&\left\langle\Phi r -\Phi r^\star, \Pi \mathfrak{F}\Phi r -\Phi r\right\rangle    
\\&=\left\langle\Phi r -\Phi r^\star, (\Pi \mathfrak{F}\Phi r -\Phi r^\star)+ \Phi r^\star -\Phi r\right\rangle
\\&\leq\left\|\Phi r -\Phi r^\star\right\|\left\|\Pi \mathfrak{F}\Phi r -\Phi r^\star\right\|- \left\|\Phi r^\star -\Phi r\right\|^2
\\&\leq(\gamma -1)\left\|\Phi r^\star -\Phi r\right\|^2,
\end{align*}
which is negative since $\gamma<1$ which completes the proof of part i).

The proof of part ii) is straightforward since we readily observe that
\begin{align*}
    \boldsymbol{\Xi}_{2,k}(r^\star)= \left\langle\phi_l, \mathfrak{F}\Phi r^\star-\Phi r\right\rangle= \left\langle\phi_l, \Pi \mathfrak{F}\Phi r^\star-\Phi r\right\rangle=0,
\end{align*}
as required and from which we deduce the result.
\end{proof}
To prove the theorem, we make use of a special case of the following result:

\begin{theorem}[Th. 17, p. 239 in \cite{benveniste2012adaptive}] \label{theorem:stoch.approx.}
Consider a stochastic process $r_t:\mathbb{R}\times\{\infty\}\times\Omega\to\mathbb{R}^k$ which takes an initial value $r_0$ and evolves according to the following:
\begin{align}
    r_{t+1}=r_t+\alpha \Xi_2(s_t,r_t),
\end{align}
for some function $s:\mathbb{R}^{2d}\times\mathbb{R}^k\to\mathbb{R}^k$ and where the following statements hold:
\begin{enumerate}
    \item $\{s_t|t=0,1,\ldots\}$ is a stationary, ergodic Markov process taking values in $\mathbb{R}^{2d}$
    \item For any positive scalar $q$, there exists a scalar $\mu_q$ such that $\mathbb{E}\left[1+\|s_t\|^q|s\equiv s_0\right]\leq \mu_q\left(1+\|s\|^q\right)$
    \item The step size sequence satisfies the Robbins-Monro conditions, that is $\sum_{t=0}^\infty\alpha_t=\infty$ and $\sum_{t=0}^\infty\alpha^2_t<\infty$
    \item There exists scalars $c$ and $q$ such that $    \|\Xi_2(w,r)\|
        \leq c\left(1+\|w\|^q\right)(1+\|r\|)$
    \item There exists scalars $c$ and $q$ such that $
        \sum_{t=0}^\infty\left\|\mathbb{E}\left[\Xi_2(w_t,r)|z_0\equiv z\right]-\mathbb{E}\left[\Xi_2(w_0,r)\right]\right\|
        \leq c\left(1+\|w\|^q\right)(1+\|r\|)$
    \item There exists a scalar $c>0$ such that $
        \left\|\mathbb{E}[\Xi_2(w_0,r)]-\mathbb{E}[\Xi_2(w_0,\bar{r})]\right\|\leq c\|r-\bar{r}\| $
    \item There exists scalars $c>0$ and $q>0$ such that $
        \sum_{t=0}^\infty\left\|\mathbb{E}\left[\Xi_2(w_t,r)|w_0\equiv w\right]-\mathbb{E}\left[\Xi_2(w_0,\bar{r})\right]\right\|
        \leq c\|r-\bar{r}\|\left(1+\|w\|^q\right) $
    \item There exists some $r^\star\in\mathbb{R}^k$ such that $\boldsymbol{\Xi}_2(r)(r-r^\star)<0$ for all $r \neq r^\star$ and $\bar{s}(r^\star)=0$. 
\end{enumerate}
Then $r_t$ converges to $r^\star$ almost surely.
\end{theorem}

In order to apply the Theorem \ref{theorem:stoch.approx.}, we show that conditions 1 - 7 are satisfied.

Conditions 1-2 are true by assumption while condition 3 can be made true by choice of the learning rates. Therefore it remains to verify conditions 4-7 are met.   

To prove 4, we observe that
\begin{align*}
\left\|\Xi_2(w,r)\right\|
&=\left\|\phi(s)\left(L(z)+\gamma\max\left\{(\Phi r) (z'),\mathcal{M}^{\boldsymbol{\pi}}\Phi (z')\right\}-(\Phi r)(z)\right)\right\|
\\&\leq\left\|\phi(z)\right\|\left\|L(z)+\gamma\left(\left\|\phi(z')\right\|\|r\|+\mathcal{M}^{\boldsymbol{\pi}}\Phi (z')\right)\right\|+\left\|\phi(z)\right\|\|r\|
\\&\leq\left\|\phi(z)\right\|\left(\|L(z)\|+\gamma\|\mathcal{M}^{\boldsymbol{\pi}}\Phi (z')\|\right)+\left\|\phi(z)\right\|\left(\gamma\left\|\phi(z')\right\|+\left\|\phi(z)\right\|\right)\|r\|.
\end{align*}
Now using the definition of $\mathcal{M}$, we readily observe that $\|\mathcal{M}^{\boldsymbol{\pi}}\Phi (z')\|\leq \| \cR\|+\gamma\|\mathcal{P}^\pi_{s's_t}\Phi\|\leq \| \cR\|+\gamma\|\Phi\|$ using the non-expansiveness of $P$.

Hence, we lastly deduce that
\begin{align*}
\left\|\Xi_2(w,r)\right\|
&\leq\left\|\phi(z)\right\|\left(\|L(z)\|+\gamma\|\mathcal{M}^{\boldsymbol{\pi}}\Phi (z')\|\right)+\left\|\phi(z)\right\|\left(\gamma\left\|\phi(z')\right\|+\left\|\phi(z)\right\|\right)\|r\|
\\&\leq\left\|\phi(z)\right\|\left(\|L(z)\|+\gamma\| \cR\|+\gamma\|\psi\|\right)+\left\|\phi(z)\right\|\left(\gamma\left\|\phi(z')\right\|+\left\|\phi(z)\right\|\right)\|r\|,
\end{align*}
we then easily deduce the result using the boundedness of $\phi,\cR$ and $\psi$.

Now we observe the following Lipschitz condition on $\Xi_2$:
\begin{align*}
&\left\|\Xi_2(w,r)-\Xi_2(w,\bar{r})\right\|
\\&=\left\|\phi(z)\left(\gamma\max\left\{(\Phi r)(z'),\mathcal{M}^{\boldsymbol{\pi}}\Phi(z')\right\}-\gamma\max\left\{(\Phi \bar{r})(z'),\mathcal{M}^{\boldsymbol{\pi}}\Phi(z')\right\}\right)-\left((\Phi r)(z)-\Phi\bar{r}(z)\right)\right\|
\\&\leq\gamma\left\|\phi(z)\right\|\left\|\max\left\{\phi'(z') r,\mathcal{M}^{\boldsymbol{\pi}}\Phi'(z')\right\}-\max\left\{(\phi'(z') \bar{r}),\mathcal{M}^{\boldsymbol{\pi}}\Phi'(z')\right\}\right\|+\left\|\phi(z)\right\|\left\|\phi'(z) r-\phi(z)\bar{r}\right\|
\\&\leq\gamma\left\|\phi(z)\right\|\left\|\phi'(z') r-\phi'(z') \bar{r}\right\|+\left\|\phi(z)\right\|\left\|\phi'(z) r-\phi'(z)\bar{r}\right\|
\\&\leq \left\|\phi(z)\right\|\left(\left\|\phi(z)\right\|+ \gamma\left\|\phi(z)\right\|\left\|\phi'(z') -\phi'(z') \right\|\right)\left\|r-\bar{r}\right\|
\\&\leq c\left\|r-\bar{r}\right\|,
\end{align*}
using Cauchy-Schwarz inequality and  that for any scalars $a,b,c$ we have that $
    \left|\max\{a,b\}-\max\{b,c\}\right|\leq \left|a-c\right|$.
    
Using Assumptions 3 and 4, we therefore deduce that
\begin{align}
\sum_{t=0}^\infty\left\|\mathbb{E}\left[\Xi_2(w,r)-\Xi_2(w,\bar{r})|w_0=w\right]-\mathbb{E}\left[\Xi_2(w_0,r)-\Xi_2(w_0,\bar{r})\right\|\right]\leq c\left\|r-\bar{r}\right\|(1+\left\|w\right\|^l).
\end{align}

Part 2 is assured by Lemma \ref{projection_F_contraction_lemma} while Part 4 is assured by Lemma \ref{value_difference_Q_difference} and lastly Part 8 is assured by Lemma \ref{iteratation_property_lemma}.
\end{proof}

To complete the proof of Theorem \ref{convergence_theorem}, we make use of Theorem 1.1. in \cite{borkar1997stochastic} in which case we readily verify that with the appropriate choices of timesteps the Theorem is readily satisfied.

\section{Alternative uncertainty measures}\label{app:uncertainty-measures}
$\bullet$ \textbf{Model-Based Ensemble Disagreement.} By employing an ensemble of dynamics models, $\{\mathcal{M}_{1}, \dots, \mathcal{M}_{E}\}$, where each $\mathcal{M}_{e}\in\mathcal{F}$ for $e=1, \dots, E$. Model training entails independent training of each of the model in the ensemble with the identical objectives (e.g., minimising the L2 distance between the predicted and the ground-truth next states). The uncertainty about a state-action pair $(s, a)$, can be quantified as the predictive ensemble disagreement: 
    \begin{equation}
        L(s, a) = \frac{1}{E-1}\sum_{e}(\mathcal{M}_{e}(s, a) - \mu(s, a))
        \label{eq: ensemble-var}
    \end{equation}
    where $\mu(s, a) = \frac{1}{E}\sum_{e}\mathcal{M}_{e}(s, a)$ is the empirical mean of the ensemble predictions. This approach has a information-theoretic interpretation such that through training, the mutual information between the dynamics model parameters and next-state is maximised, hence relating the epistemic uncertainty with the information-theoretic framework. 
    
    $\bullet$ \textbf{Integrating control into dynamics modelling with LSSM.} Consider we embed the dynamics modelling problem into a sequential modelling problem using latent state-space models (LSSM), using amortised inference, we are able to achieve fast inference and learning of the probabilistic graphical model. We could additionally incorporate action into the LSSM as a global factor that (potentially) influences both the latent and observable codes. For instance, the generative process could be modelled as:
    \begin{align*}
        &p(\mathbf{x}_{1:T}, \mathbf{z}_{1:T}, \mathbf{a}_{1:T-1}) = p(z_{1})p(x_{1}|z_{1}) \cdot 
        \\&\cdot \prod_{t=2}^{T}p(z_{t}|z_{t-1}, a_{t-1})p(x_{t}|z_{t})p(a_{t-1}|z_{t-1}, x_{t-1}).
    \end{align*}
    We could easily train an LSSM by maximising the variational lower bound utilising amortised inference. In the meantime, we could quantify the model uncertainty about the state-action pair $(z_{t}, a_{t})$ in terms of the variance of the latent predictive distributions. By random trajectory-sampling (multiple particles), we target regions of the action space that maximises the predictive variance (assuming Gaussian for now). This could be achieved by importance-weighting on the computation of the marginal variance. Hence in this case we use the following uncertainty instantiation:
    \begin{equation}
        L(s, a) = \mathbb{V}(s'|s, a).
        \label{eq: lssm-var}
    \end{equation}
    We could consider \eqref{eq: lssm-var} as a parametric generalisation of \eqref{eq: ensemble-var} (despite the fact that in the model-ensemble method, the action is taken as an external input instead of a random variable as in the LSSM method).
    
\section{Implementation Details}
\label{sec: implementation_details}
For all implemented deep RL agents (SEREN-DQN, SEREN-SAC, SEREN-TD3), we use MLPs as the function approximator, with Adam optimiser~\cite{kingma2014adam}. 
We show the implementation details of the SEREN-SAC agent in Table~\ref{table: implementation_details} that are used in all studied MuJoCo environments. The implementation of SEREN-TD3 agent is mostly similar to that of SEREN-SAC, and only differs in the learning rates ($5\times 10^{-4}$ for {\fontfamily{cmss}\selectfont Exploiter}, {\fontfamily{cmss}\selectfont Explorer} and {\fontfamily{cmss}\selectfont Switcher}, and intervention cost ($\beta$) is changed from being $10$ to being $50$).

\begin{table*}[t]
\begin{center}
\begin{small}
\begin{sc}
\begin{tabular}{lcccr}
\toprule
Component & Attribute & Value \\
\midrule
{\fontfamily{cmss}\selectfont Exploiter} & critic MLP hidden layer dimensions & $[256, 256]$ \\
 & critic MLP activation function & ReLU\\
 & actor MLP hidden layer dimensions & $[256, 256]$ \\
 & actor MLP activation function & ReLU \\
 & learning rate & $5\times 10^{-4}$ \\
 & replay buffer size & $2\times 10^{5}$ \\
 & batch size & 256 \\
 & number of critic ensemble & 5 \\
 & discounting factor & 0.99 \\
 \midrule
{\fontfamily{cmss}\selectfont Explorer} & critic MLP hidden layer dimensions & $[256, 256]$ \\
  & critic MLP activation function & ReLU\\
 & actor MLP hidden layer dimensions & $[256, 256]$ \\
 & actor MLP activation function & ReLU \\
 & learning rate & $3\times 10^{-4}$ \\
 & replay buffer size & $2\times 10^{5}$ \\
 & discounting factor & 0.05 \\
\midrule
{\fontfamily{cmss}\selectfont Switcher} & critic MLP hidden layer dimensions & $[64, 64]$ \\ 
                 & critic MLP activation function & ReLU \\ 
                 & actor MLP hidden layer dimensions & $[64, 64]$\\
                 & actor MLP activation function & ReLU \\ 
                 & learning rate & $3\times 10^{-4}$\\
                 & replay buffer size & $2\times 10^{5}$ \\ 
\midrule
SEREN & {\fontfamily{cmss}\selectfont Switcher} intervention cost & $10.0$ \\ 
             & number of initial exploration steps & $10000$ \\ 
             & frequency of training {\fontfamily{cmss}\selectfont Exploiter} & 8 \\
             & frequency of training {\fontfamily{cmss}\selectfont Exploiter} and {\fontfamily{cmss}\selectfont Switcher} & 4 \\
             & proportion of batch data for training ensemble components & $0.8$\\
\bottomrule
\end{tabular}
\end{sc}
\end{small}
\end{center}
\caption{Implementation specifications for SEREN-SAC for all MuJoCo environments considered.}
\label{table: implementation_details}
\end{table*}

\section{Further Experimental Results}
\label{sec: additional_experiment}

\begin{figure}
     \centering
     \includegraphics[width=\linewidth]{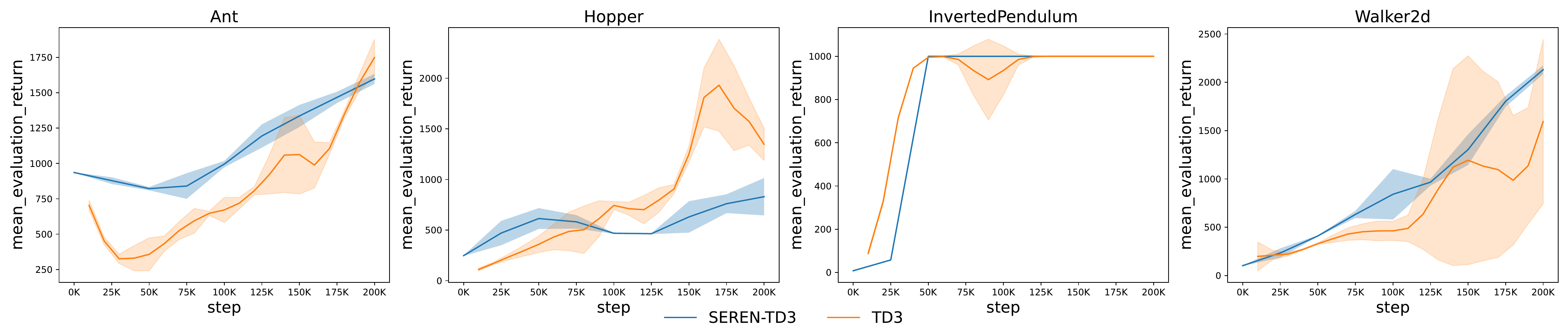}
     \caption{SEREN-TD3 on selected MuJoCo environments.}
     \label{fig: td3_mujoco}
 \end{figure}

We proposed SEREN as a general framework that can be readily combined with any reinforcement learning agent to promote optimal balancing between exploration and exploitation.
In Figure~\ref{fig: td3_mujoco} we show the performance of SEREN-TD3, and the comparison with standard TD3 on selected MuJoCo tasks~\cite{fujimoto2018addressing}. We observe that SEREN-TD3 outperforms or achieves similar performance as standard TD3 on $3$ out of the $4$ presented tasks.

 \end{document}